\documentclass[letterpaper]{article} 
\usepackage{aaai2026}  
\usepackage{times}  
\usepackage{helvet}  
\usepackage{courier}  
\usepackage[hyphens]{url}  
\usepackage{graphicx} 
\usepackage{natbib}  
\usepackage{caption} 
\usepackage{amssymb}
\usepackage{booktabs}
\usepackage{pifont}
\usepackage{amsmath}
\usepackage{multirow}
\usepackage{makecell}
\usepackage{algorithm}
\usepackage{algorithmic}

\usepackage{newfloat}
\usepackage{listings}
\DeclareCaptionStyle{ruled}{labelfont=normalfont,labelsep=colon,strut=off} 
\floatstyle{ruled}
\newfloat{listing}{tb}{lst}{}
\floatname{listing}{Listing}
\title{AerialMind: Towards Referring Multi-Object Tracking in UAV Scenarios}
\author{
    Chenglizhao Chen\textsuperscript{\rm 1,2}, Shaofeng Liang\textsuperscript{\rm 1,2}, Runwei Guan\textsuperscript{\rm 3}
    \thanks{Corresponding author: runwayrwguan@hkust-gz.edu.cn}, Xiaolou Sun\textsuperscript{\rm 4}, Haocheng Zhao\textsuperscript{\rm 5}, \\
    Haiyun Jiang\textsuperscript{\rm 6}, Tao Huang\textsuperscript{\rm 7}, Henghui Ding\textsuperscript{\rm 8}, Qing-Long Han\textsuperscript{\rm 9}
}

\affiliations{
    \textsuperscript{\rm 1}Qingdao Institute of Software, College of Computer Science and Technology, China University of Petroleum (East China)\\
    \textsuperscript{\rm 2}Shandong Key Laboratory of Intelligent Oil \& Gas Industrial Software\\
    \textsuperscript{\rm 3}Thrust of Artificial Intelligence, The Hong Kong University of Science and Technology (Guangzhou) \\
    \textsuperscript{\rm 4}Purple Mountain Laboratories \\
    \textsuperscript{\rm 5}School of Advanced Technology, Xi'an Jiaotong-Liverpool University\\   \textsuperscript{\rm 6}School of Automation and Intelligent Sensing, Shanghai Jiao Tong University \\
    \textsuperscript{\rm 7}College of Science and Engineering, James Cook University  \\
    \textsuperscript{\rm 8}Institute of Big Data, College of Computer Science and Artificial Intelligence, Fudan University \\ \textsuperscript{\rm 9}School of Engineering, Swinburne University of Technology, Melbourne\\

}

\usepackage{bibentry}

\begin{document}

\maketitle

\begin{abstract}
Referring Multi-Object Tracking (RMOT) aims to achieve precise object detection and tracking through natural language instructions, representing a fundamental capability for intelligent robotic systems. However, current RMOT research remains mostly confined to ground-level scenarios, which constrains their ability to capture broad-scale scene contexts and perform comprehensive tracking and path planning. In contrast, Unmanned Aerial Vehicles (UAVs) leverage their expansive aerial perspectives and superior maneuverability to enable wide-area surveillance. Moreover, UAVs have emerged as critical platforms for Embodied  Intelligence, which has given rise to an unprecedented demand for intelligent aerial systems capable of natural language interaction. 
To this end, we introduce AerialMind, the first large-scale RMOT benchmark in UAV scenarios, which aims to bridge this research gap.  To facilitate its construction, we develop an innovative semi-automated collaborative agent-based labeling assistant (COALA) framework that significantly reduces labor costs while maintaining annotation quality. Furthermore, we propose HawkEyeTrack (HETrack), a novel method that collaboratively enhances vision-language representation learning and improves the perception of UAV scenarios. Comprehensive experiments validated the challenging nature of our dataset and the effectiveness of our method.
\end{abstract}

\begin{links}
    \link{Datasets}{https://github.com/shawnliang420/AerialMind}
\end{links}

\section{Introduction}
Referring Multi-Object Tracking (RMOT)~\cite{refer-kitti,refer-kittiv2} aims to achieve precise detection and tracking of specified targets in video sequences through language instructions. It realizes a fundamental paradigm shift from passive perception to active understanding. Although significant progress~\cite{refer-dance,refer-bdd,mls,refer-kitti,delving} has been achieved, it is almost entirely confined to ground-level scenarios. It constrains their ability to capture broad-scale scene contexts and perform comprehensive tracking and path planning. In contrast, Unmanned Aerial Vehicles (UAVs) leverage expansive aerial perspectives and superior maneuverability to enable wide-area surveillance capabilities unattainable by ground-based systems. As critical platforms for Embodied AI~\cite{wang2025uav}, UAVs drive unprecedented demand for intelligent aerial systems with natural language interaction capabilities. However, current RMOT research lacks sufficient exploration of challenging aerial scenarios, resulting in limited real-world applicability and hindering the realization of truly aerial intelligence.
\begin{figure}[t]
    \includegraphics[width=0.99\linewidth]{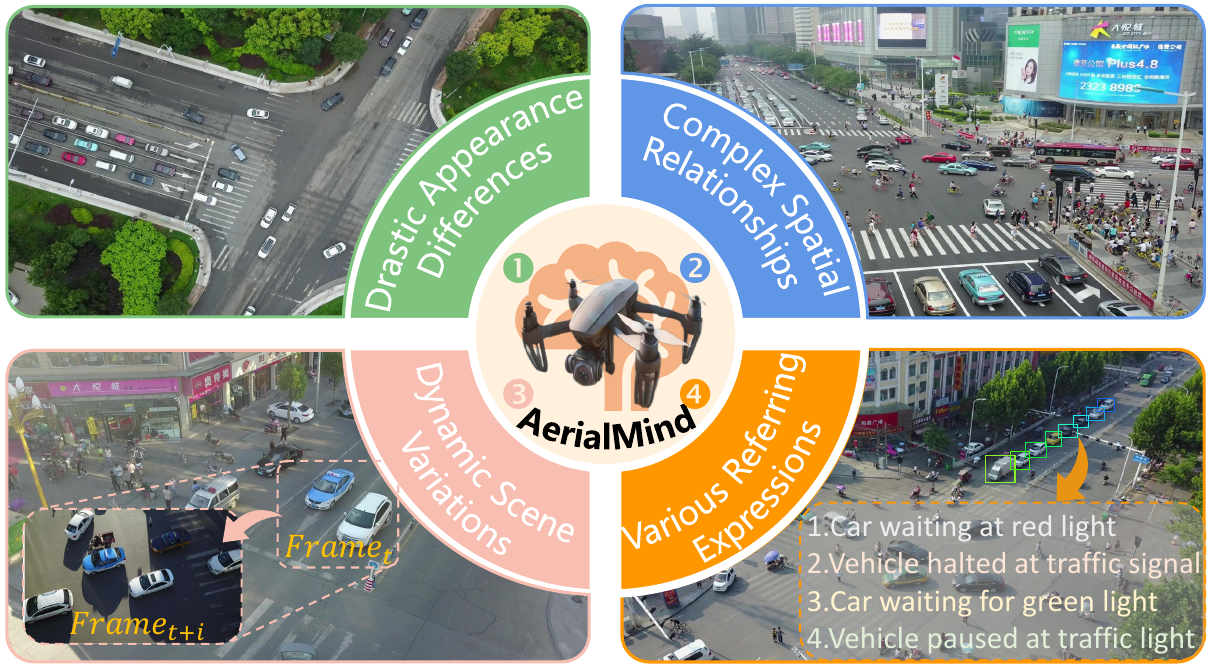}
    \caption{Overview of the challenges in AerialMind dataset.}
    \label{fig:motivation}
\end{figure}

To this end, we construct the first large-scale referring multi-object tracking dataset \textbf{AerialMind} for UAV scenarios. The dataset is extended based on VisDrone~\cite{visdrone} and UAVDT~\cite{UAVDT}, covering multiple flight altitudes, environmental conditions, and target categories. As shown in Figure~\ref{fig:motivation}, AerialMind brings unprecedented challenges: \ding{182} Drastic Appearance Difference: Changes in flight altitude and viewpoints cause dramatic differences in object appearance; \ding{183} Complex Spatial Relationships: Object relationships under aerial view perspectives are more intricate; \ding{184} Dynamic Scene Variations: The high maneuverability of UAVs brings continuously changing scenes and illumination conditions; \ding{185} Various Referring Expressions: Spatial, motion states, and object descriptions in UAV scenarios exhibit richer semantic complexity. To facilitate profound and quantitative analysis, we also pioneer frame-by-frame attribute annotations in the RMOT field.

\begin{table*}[t]
    \centering
    \small  
    \setlength{\tabcolsep}{3.5pt} 
    \renewcommand{\arraystretch}{1.1}  
    \begin{tabular}{@{}lcccccccccc@{}}
        \toprule[1.5pt]
        \textbf{Dataset} & \textbf{Source} & \textbf{Videos} & \textbf{Dom.} & \textbf{Reas.} & \textbf{Attr.} & \textbf{Expressions} &  \textbf{Words} & \textbf{Instance / Expression} & \textbf{Instances} & \textbf{Bbox Anno.} \\
        \midrule[1pt]
        Refer-KITTI & CVPR$_{2023}$    & 18 & \ding{55} & \ding{55} & \ding{55} & 818 & 49 & 10.7 & 8.8K & 0.36M \\
        Refer-Dance & CVPR$_{2024}$     & 65 & \ding{55} & \ding{55} & \ding{55}& 1.9K & 25 & 0.34 & 650 & 0.55M \\
        Refer-KITTI-V2& arXiv$_{2024}$  & 21 & \ding{55} & \ding{55} & \ding{55}& 9.8K & 617 & 6.7 & 65.4K & 3.06M \\
        Refer-UE-City& arXiv$_{2024}$  & 12 & \ding{55} & \ding{55} & \ding{55}& 714 & -- & 10.3 & -- & 0.55M \\
        Refer-BDD & IEEE TIM$_{2025}$       & 50 & \ding{55} & \ding{55} & \ding{55}& 4.6K & 225 & 15.3 & 70.4K & 1.50M \\
        CRTrack\nocite{refer-cross}& AAAI$_{2025}$  & 41 & \ding{51} & \ding{55} & \ding{55} & 344 & 43 & -- & -- & 0.79M \\
        LaMOT*& IEEE ICRA$_{2025}$        & 62& \ding{55} & \ding{55} & \ding{55} & 145 & 9 & \textbf{54.6} & 7.9K & 1.2M \\
        AerialMind & Ours  & \textbf{93} & \ding{51} & \ding{51} & \ding{51} & \textbf{24.6K} &  \textbf{1.2K} & 11.9 & \textbf{293.1K} & \textbf{46.14M} \\
        \bottomrule[1.5pt]
    \end{tabular}
    \caption{Comparison of referring multi-object tracking datasets. The Dom. represents cross-domain scenarios, Reas. denotes complex reasoning expressions,  and Attr. indicates attribute annotation. LaMOT*~\cite{lamot} represents the UAV subset.}
    \label{tab:dataset_stats}
\end{table*}

To efficiently construct AerialMind, we develop a novel semi-automated annotation framework, namely \textbf {CO}llaborative \textbf{A}gent-based \textbf{L}abeling \textbf{A}ssistant (COALA).  It aims to reduce annotation costs through intelligent processes while effectively avoiding subjective biases in manual annotation. Specifically, COALA adopts a multi-stage annotation mechanism: First, it utilizes large language models (LLMs) to intelligently parse UAV scenarios; Then, the system automatically records targeted objects by annotators simply click and define the temporal boundaries of referring events, and associates corresponding description items; Subsequently, it performs cross-modal logical reasoning on static frames and trajectory data to validate annotation quality. Finally, it leverages the generative capabilities of LLMs to expand and generate more semantically rich expressions. 

Furthermore, we propose a novel method called HawkEyeTrack (HETrack). It innovatively introduces the Co-evolutionary Fusion Encoder (CFE) that enables a co-evolutionary refinement of vision and language representations and incorporates a targeted Scale Adaptive Contextual Refinement (SACR) module to significantly enhance the perception of UAV scenarios. Comprehensive experiments on AerialMind validate the challenging nature of the benchmark and demonstrate the effectiveness of HETrack. 

In summary, our contributions are listed as follows:
\begin{enumerate}

\begin{figure*}[t]
    \centering
    \includegraphics[width=0.99\linewidth]{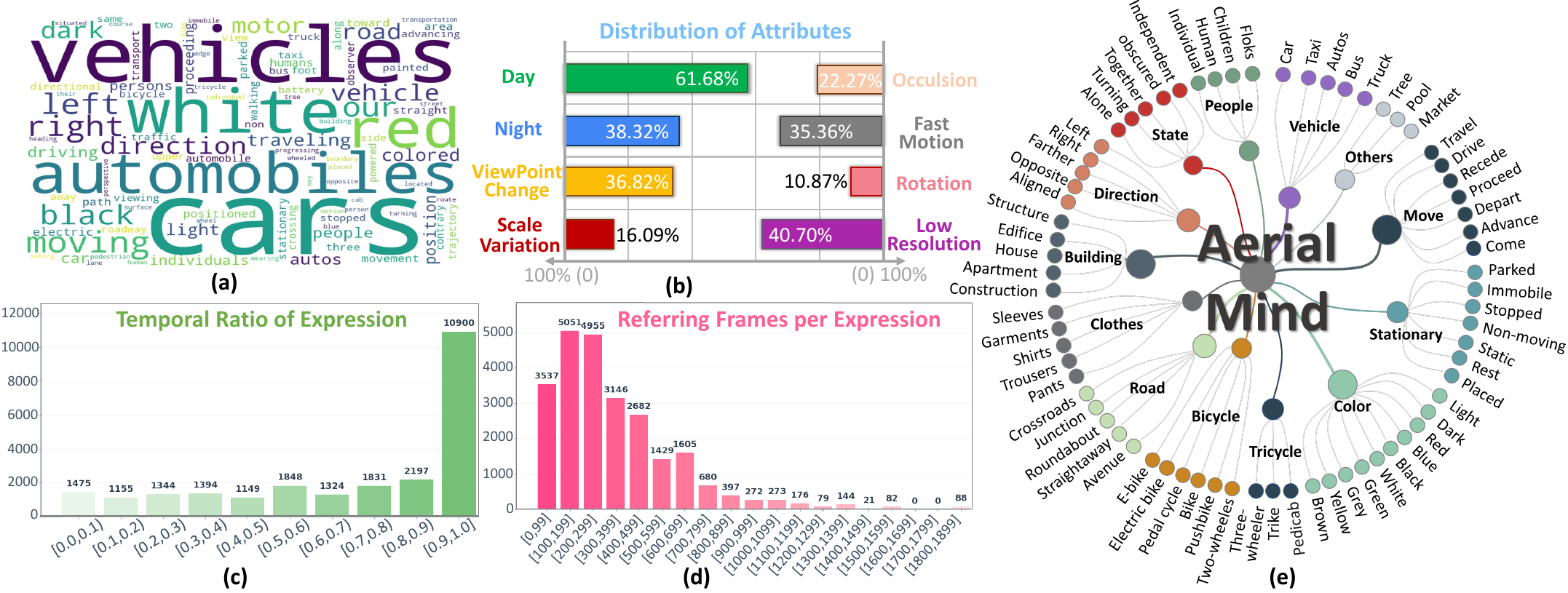}
    \caption{Overview of the AerialMind dataset statistics. It shows the distribution and diversity of (a) vocabulary, (b) challenging attributes, (c-d) temporal characteristics, and (e) semantic concepts.}
    \label{fig:vis_analysis}
\end{figure*}
\item \textbf{AerialMind benchmark dataset}: We construct the first large-scale referring multi-object tracking benchmark dataset for Unmanned Aerial Vehicle (UAV)  scenarios. It introduces new challenges for RMOT research.

\item \textbf{COALA annotation framework}: An innovative semi-automated annotation framework that adopts multi-stage agent collaborative mechanisms, significantly reducing manual costs while ensuring high-quality annotations.

\item \textbf{HETrack method}:  It integrates the co-evolutionary refinement of vision and language representations and scale adaptive contextual refinement, achieving excellent performance on our AerialMind dataset.
\end{enumerate}

\section{Related Works}
\subsection{Referring Understanding  Datasets}
Referring to understanding tasks~\cite{ding2022vlt,ding2023mose,ding2025mevis,MOSEv2,guan2024nanomvg,guan2025talk2radar,guan2025yu}, which aim to localize specific regions in images or videos through natural language expressions. Early dataset construction work mainly focused on static image scenarios, such as the RefCOCO~\cite{refcoco} series datasets. Subsequently, researchers gradually extended referring understanding to temporal video domains, successively proposing video referring segmentation datasets such as Refer-DAVIS$_{17}$~\cite{refer-davis} and Refer-Youtube-VOS~\cite{refer-youtube}.  Wu et al. ~\cite{refer-kitti} first proposed the referring multi-object tracking task. Researchers further extended this work, proposing larger-scale Refer-KITTI-V2~\cite{refer-kittiv2}, Refer-BDD~\cite{refer-bdd} and ReaMOT~\cite{reamot} datasets. They mainly focus on specific ground perspectives, lacking sufficient consideration for the unique challenges of aerial platforms such as UAVs.
Recently,  Researchers~\cite{refdrone,liu2025aerialvg} constructed the  UAV referring expression detection datasets, validating the feasibility of referring understanding from aerial perspectives. However, they concentrate on single-frame detection tasks, lacking in-depth exploration that requires long-term temporal modeling and complex language understanding.

\subsection{Referring Understanding Methods}
Early referring understanding methods~\cite{refer-davis,early2} mostly adopted two-stage strategies~\cite{early4}, which rely heavily on candidate region quality and have low computational efficiency. Currently, end-to-end methods~\cite{onestage1,onestage2} have gradually become mainstream. These methods achieve visual-language fusion through designing sophisticated mechanisms~\cite{onestage3,onestage4,video1,video2,video3}. 
For referring to multi-object tracking, TransRMOT~\cite{refer-kitti} first proposed an end-to-end solution based on the Transformer. TempRMOT~\cite{refer-kittiv2} further introduced temporal enhancement modules, improving the temporal consistency of tracking. Although these methods~\cite{refer-dance,refer-bdd} have achieved significant progress in ground scenarios, they still show inadequate adaptability when facing unique UAV challenges.

\section{Benchmark}
We construct the first large-scale referring multi-object tracking benchmark \textbf{AerialMind} for unmanned aerial vehicle (UAV) scenarios.   We demonstrate the core challenges presented by AerialMind in Figure~\ref{fig:motivation} and provide detailed data statistical analysis in Figure~\ref{fig:vis_analysis}. 
\subsection{Dataset Features and Statistics}
As shown in Table~\ref{tab:dataset_stats}, AerialMind contains 93 video sequences, totaling 24.6K referring expressions, associated with 293.1K object instances and up to 46.14M bounding box annotations. In comparison, even the larger-scale Refer-KITTI-V2 has only 9.8K expressions, less than half of AerialMind. More importantly, AerialMind systematically covers cross-domain scenarios and complex referring expressions (including 752 no-target expressions and 458 reasoning expressions) and fine-grained attribute annotations, greatly enhancing the comprehensive challenge of the task. 
The word clouds and semantic concepts are as shown in Figure~\ref{fig:vis_analysis}-a \& e, demonstrating the rich linguistic diversity and semantic breadth of our dataset. The temporal ratio distribution of referring expressions in videos (Figure~\ref{fig:vis_analysis}-c) is broad and balanced, meaning referring events may occur or end at any point in the video. The representative frame count distribution (see Figure~\ref{fig:vis_analysis}-d) exhibits obvious long-tail characteristics, with many long-term referring events spanning hundreds of frames. This presents significant challenges for models' temporal event localization abilities.

To promote deeper and more refined diagnostic analysis of model performance, we introduce a novel attribute-based evaluation, which is the first exploration in the RMOT field.  We frame-by-frame annotate eight challenge attributes in the test set: illumination conditions (day/night), viewpoint change, scale variation, occlusion, fast motion, camera rotation, and low resolution, as shown in Figure~\ref{fig:vis_analysis}-b. We introduced new metrics, namely HOTA$_S$ and HOTA$_M$, to evaluate the model's capability in addressing scene-induced and motion-induced challenges, respectively.

\subsection{Collaborative Agent-based Labeling Assistant}
\begin{figure*}[t]
    \includegraphics[width=\linewidth]{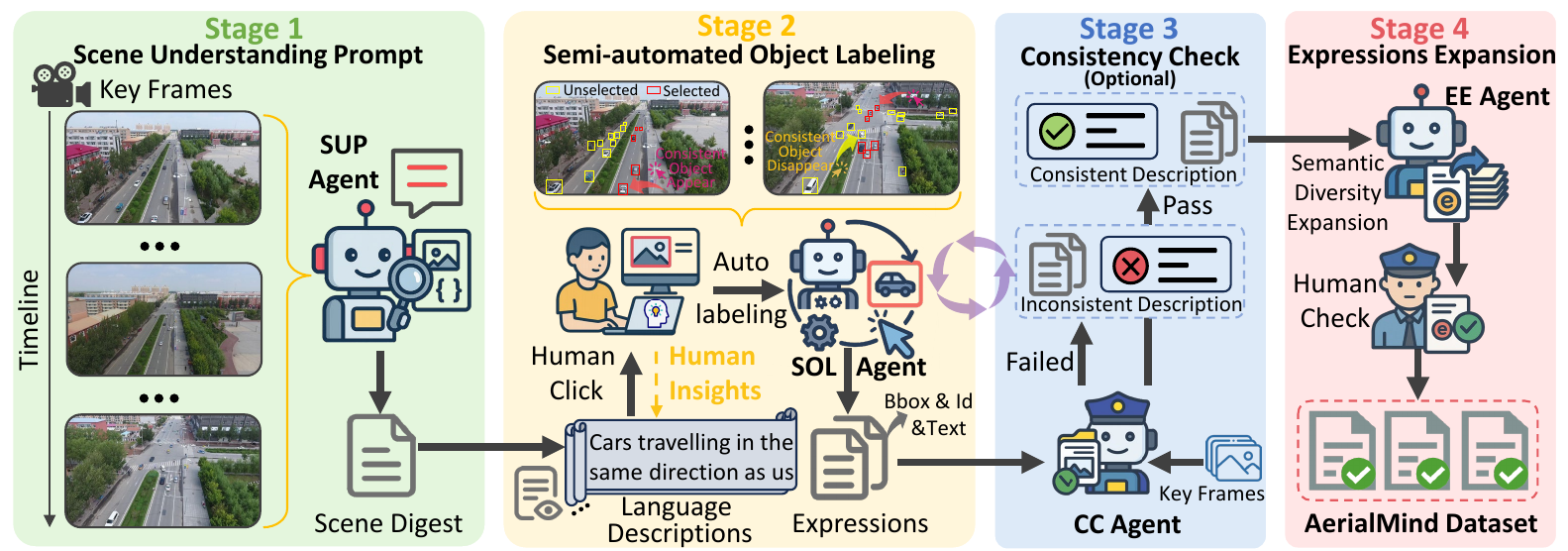}
    \caption{Overview of the four-stage annotation process in the COALA framework. This framework efficiently constructs the AerialMind dataset through multi-agent collaboration and human-computer interaction.}
    \label{fig:anno}
\end{figure*}
Traditional annotation pipelines for referring expressions are labor-intensive and time-costly. Consequently, we introduce the \textbf{CO}llaborative \textbf{A}gent-based \textbf{L}abeling \textbf{A}ssistant (\textbf{COALA}), a novel semi-automated framework designed to augment the traditional annotation pipeline through the LLM agent. 
It explores a sustainable and scalable next-generation visual annotation paradigm that achieves a balance between cost, efficiency, quality, and diversity. 

\textbf{1. Scene Understanding and Prompt Generation}: To alleviate the high cognitive load and time costs caused by tedious manual video review, we first utilize the few-shot visual question answering (VQA) capabilities of large language models (LLMs) to guide the annotation process. For each video, we design a Scene Understanding Prompt Agent (SUP-Agent). This agent takes key frames of the video as input and automatically generates high-level semantic digests of the scene. It covers descriptions of key objects, attributes, and spatial layouts and includes a series of templated scene prompts, providing a structured starting point for subsequent annotation tasks, as shown in Figure~\ref{fig:anno}-Stage 1.

\textbf{2. Semi-automated Object Labeling}: Figure~\ref{fig:anno}-Stage 2 shows that we introduce a human-machine collaborative workflow, whose core is the Semi-automated Object Labeling Agent (SOL-Agent). The human annotator first reads the digest from Stage 1 to gain a comprehensive understanding of the video's context. Subsequently, the annotator actively selects a language description from the templated scene prompts. With the chosen text as guidance, annotators only need to perform two clicks on a target instance to define the complete temporal interval (i.e., the start and end points) where it matches the description. The SOL-Agent then tracks and associates its corresponding bounding box trajectory frame by frame based on the existing detection boxes in the video. This ``click-to-define" interaction paradigm liberates annotators from tedious frame-by-frame operations, allowing them to focus on high-level semantic judgment and temporal boundary definition, greatly improving annotation efficiency and consistency. More importantly, when human experts identify more complex or subtle interactions that are not covered by preset scene prompts during the annotation process. They can directly create new, more precise linguistic descriptions instantly, ensuring comprehensive coverage of complex real-world scenarios.

\textbf{3. Consistency Check}: To ensure the highest quality of annotations and lay the foundation for future fully automated processes, we introduce an optional but crucial Consistency Check Agent (CC-Agent), as illustrated in Figure~\ref{fig:anno}-Stage 3. Its core innovation is to perform cross-modal spatio-temporal logical reasoning by analyzing a comprehensive data package based on LLM. Specifically, the CC-Agent validates the matching degree between visual features, language descriptions, and the motion patterns inferred from trajectory data (such as velocity and directional changes). Annotations that fail to pass validation will be returned for correction. This stage is designated as optional, primarily to balance its significant cost against the already high fidelity of the preceding annotations. However, it serves as the foundation for a fully automated annotation in the feature.

\textbf{4. Expression Expansion}: In the final stage, we design an Expression Expansion Agent (EE-Agent) that plays the role of a ``linguist" (see Figure~\ref{fig:anno}-Stage 4). This agent takes validated expressions as ``semantic seeds" and is prompted to generate multiple new expressions that differ in syntax and vocabulary but are semantically equivalent. This step greatly enriches the linguistic diversity of the dataset. Finally, all machine-generated expressions enter a final human verification process to thoroughly filter out any potential errors or ``hallucinations" introduced by LLMs.

\section{Method}
\begin{figure*}[t]
    \includegraphics[width=0.99\linewidth]{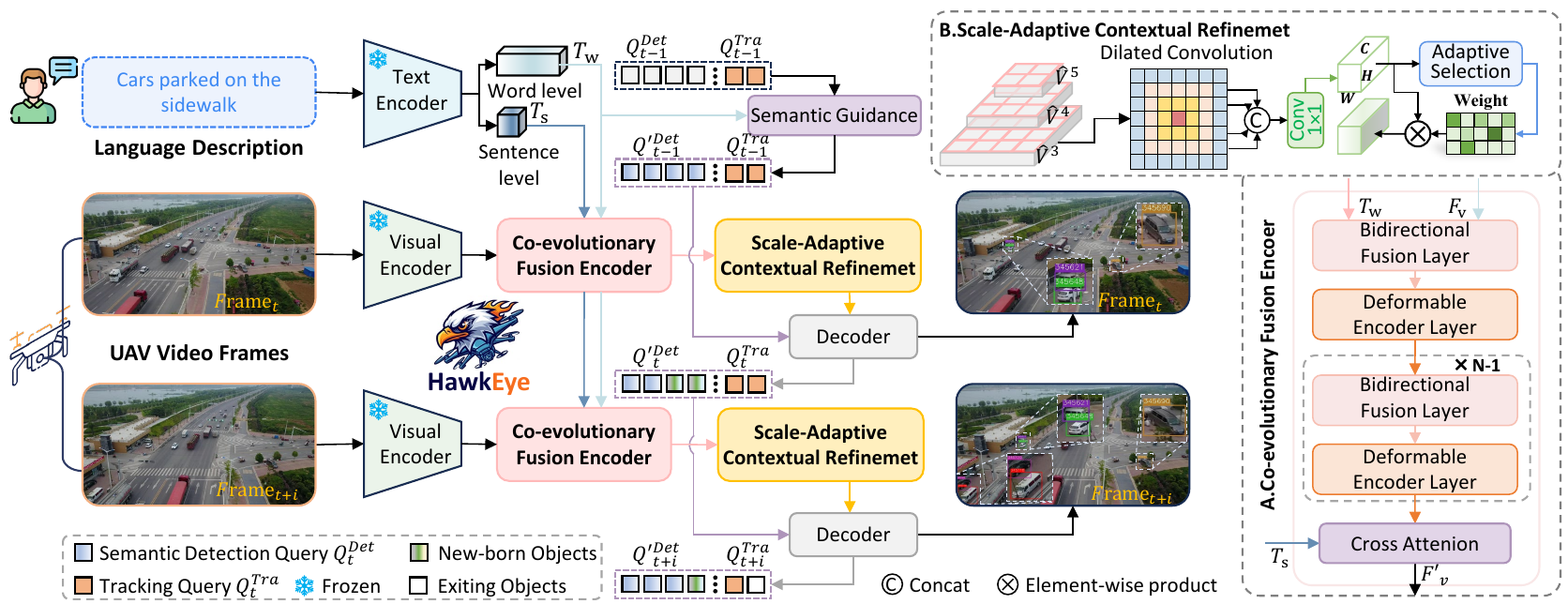}
    \caption{Overview of the HawkEyeTrack. Our key innovations include the Co-evolutionary Fusion Encoder for synergistic vision-language alignment and Scale-Adaptive Contextual Refinement for enhancing the
 perception of UAV scenarios.}
    \label{fig:pipe}
\end{figure*}

In this work, we propose a novel framework named HawkEyeTrack (HETrack) for robust referring tracking. We introduce two key innovations: a Co-evolutionary Fusion Encoder that enables collaborative refinement of vision and language representations, and a  Scale Adaptive Contextual Refinement module to significantly enhance the perception of UAV scenarios, as shown in Figure~\ref{fig:pipe}. 
\subsection{Co-evolutionary Fusion Encoder}
In language-guided visual perception tasks, achieving efficient Cross-Modal Representation Alignment~\cite{chen2024sato,chen2025ant,chen2025free} is the core challenge. Existing methods mostly follow the early fusion and late fusion paradigms. Early fusion attempts to forcibly align highly abstract text with unstructured, noisy visual features at the beginning of visual encoding. It faces a huge modality gap and may cause the language signal to be progressively diluted in the subsequent encoding. Conversely,  although late fusion structures the structural visual features, it makes this process a blind exploration without language navigation, resulting in the final fusion to an inefficient ``post-hoc correction". These become increasingly prominent when facing various descriptions in AerialMind that are full of complex spatial relationships.
To this end, we propose a novel Co-evolutionary Fusion Encoder (CFE). Our key insight is: the structuring process of visual features and the guiding process of language information should not be independent stages, but rather a deeply intertwined and mutually reinforcing unified body.

Specifically, given an image frame, a visual backbone network extracts a multi-scale feature pyramid, denoted as $\mathbf{F}_V = \{ \mathbf{V}^{(l)} \}_{l=1}^{L_s}$, where $\mathbf{V}^{(l)} \in \mathbb{R}^{H_l \times W_l \times C_l}$ represents the feature map at the $l$-th level. Concurrently, the input language expression is encoded via a text encoder into two granularities: word-level features $\mathbf{T}_w \in \mathbb{R}^{L \times C}$ and a sentence-level global feature $\mathbf{T}_s \in \mathbb{R}^{1 \times C}$.
The CFE is constructed by stacking $N_e$  blocks. Each block comprises a Bidirectional Fusion Layer (BFL) and a Deformable Encoder Layer (DEL). 
The bidirectional nature of this fusion implies that visual features provide concrete anchors for linguistic concepts, while linguistic concepts offer targeted guidance for the filtering and enhancement of visual features. 
Then, the fused features $\mathbf{F'}_V^{(i)}$ are immediately processed by a DEL for efficient intra-modal spatial relationship modeling.  After $N_e$ iterations, we obtain a final visual representation, $\mathbf{F}_{\text{enc}} = \mathbf{F'}_V^{(N_e)}$. To imbue the model with a holistic grasp of the overall referring intent, we leverage the global sentence-level feature $\mathbf{T}_s$ to perform a final modulation on the co-evolved visual features $\hat{\mathbf{F}}_V$. Formally:
\begin{equation}
\begin{aligned}
    \mathbf{F'}_V^{(i)}; \mathbf{T}^{(i)}_w &= \text{BFL}^{i}(\mathbf{F}_V^{(i)}, \mathbf{T}^{(i)}_w; \theta_i)\\
     & = \mathbf{F}_V^{(i)} + \underset{\Uparrow}{\underline{\Delta\mathbf{F}_V^{(i)}}};  \mathbf{T}^{(i)}_w+\Delta\mathbf{T}^{(i)}_w, \\[-0.5em]
      &\hspace{0.23cm}
\rm \overbrace{\text{MHA}(\mathbf{F}_V^{(i)}, \mathbf{T}^{(i)}_w, \mathbf{T}^{(i)}_w)} 
\end{aligned}
\end{equation}

\begin{equation}
    \mathbf{F}_V^{(i+1)} =  \text{DEL}^{i}(\mathbf{F'}_V^{(i)}),
\end{equation}

\begin{equation}
\begin{aligned}
&\hspace{-1.65cm}
\rm \underbrace{\text{softmax}\left(\frac{(Q \mathbf{W^Q})(K \mathbf{W^K})^T}{\sqrt{d/h}}\right)(V \mathbf{W^V})}\\[-0.5em]
&\hspace{-0.25cm}
\rm \underbrace{\text{Concat}(\overset{\Downarrow} {\overline{\text{head}_1}}, \dots, \text{head}_h) W_V^O}\\[-0.5em]
\hat{\mathbf{F}}_V &= \mathbf{F}_{\text{enc}} + \overset{\Downarrow} {\overline{\text{MHA}}}(\mathbf{F}_{\text{enc}},\mathbf{\Psi}(\mathbf{T}_s),\mathbf{\Psi}(\mathbf{T}_s)) ,
\end{aligned}
\end{equation}
where $\theta_i$ is learnable parameters, MHA denotes the Multi-head Attention, $W_V^O$ represents the linear projection matrix, $\mathbf{\Psi}(\cdot)$ is a MLP projection function, $h$ is the number of  head. 

\subsection{Scale Adaptive Contextual Refinement}
A severe challenge in UAV visual perception is the performance degradation in detecting small-scale objects. Although the Deformable DETR architecture bypasses the traditional FPN, it has an inherent shortcoming. Specifically, the high-resolution feature maps, which are crucial for localizing small objects, possess a severely limited Effective Receptive Field. This results in a significant deficiency of contextual information, making it difficult for the model to distinguish small objects from complex background noise~\cite{song2022improving,wang2024windb}. To address this, we insert a lightweight yet efficient module, named the Scale-Adaptive Contextual Refinement (SACR), between the encoder and decoder, as shown in Figure~\ref{fig:pipe}-B.

Specifically, we first employ parallel atrous convolutions with multiple distinct dilation rates on the highest-resolution feature map from the $\hat{\mathbf{F}}_V = \{ \hat{\mathbf{V}}^{(l)} \}_{l=1}^{L_s} $, denoted as $V_{ac}$. It is capable of capturing rich, multi-scale contextual information without sacrificing spatial resolution. Formally:
\begin{equation}
\begin{aligned}
V_{ac}^{(3)} = \text{Concat}\left( \text{Conv}_{1 \times 1}(\hat{\mathbf{V}}^{(3)}), \{\text{DConv}_{\{r_j\}}(\hat{\mathbf{V}}^{(3)})\}_{j=1}^M \right),
\end{aligned}
\end{equation}
where $\operatorname{DConv}_{r_j}$ represents a $3 \times 3$ atrous convolution, $\{r_j\} = \{6, 12, 18\}$ denotes dilation rate. 

\begin{table*}[t]
\setlength{\tabcolsep}{4pt} 
\begin{tabular}{@{}l|cccccc|cccccc@{}} 
\toprule[1.5pt]
\multicolumn{1}{c|}{\multirow{2}{*}{\textbf{Method}}} & \multicolumn{6}{c|}{\textbf{In-domain Evaluation}} & \multicolumn{6}{c}{\textbf{Cross-domain Evaluation}} \\
\cmidrule(lr){2-7} \cmidrule(lr){8-13}  
& HOTA & DetA & AssA & HOTA$_S$ & HOTA$_M$ & LocA & HOTA & DetA & AssA & HOTA$_S$ & HOTA$_M$ & LocA \\
\midrule
MOTR-V2$_\text{CVPR 2023}$ & 19.51 & 11.57 & 33.13 & 21.67 & 19.11 & \textbf{83.80} & 21.70 & 13.85 & 34.13 & 23.85 & 24.85 & 83.43 \\
TransRMOT$_\text{CVPR 2023}$ & 23.54 & 13.18 & 42.24 & 27.21 & 24.05 & 83.47 & 26.86 & 15.21 & 47.66 & 24.47 & 25.43 & 83.65\\
TempRMOT$_\text{arXiv 2024}$ & 26.24 & 13.06 & \textbf{53.22} & 28.14 & 23.77 & 80.41 & 27.58 & 13.46 & \textbf{56.84} & 23.74& 27.67 & 83.06\\
CDRMT$_\text{INFFUS 2025}$ & 25.81 & 14.66 & 45.69 & 27.49 & 25.80 & 83.13 &  26.68 & 16.21 & 44.11 & 26.98 & 25.20 & 83.08\\
MGLT$_\text{TIM 2025}$ & 26.16 & 14.83 & 46.47 & 26.39 & 26.10& 82.44 & 27.66 & 15.18 & 50.60&  26.94 & 28.19 & 83.94\\
HETrack (Ours) & \textbf{31.46} & \textbf{21.57} & 46.23 & \textbf{34.37} & \textbf{31.12} & 82.77 & \textbf{31.60} & \textbf{21.35} & 47.10 & \textbf{27.53} & \textbf{31.93} & \textbf{83.98} \\
\bottomrule[1.5pt]
\end{tabular}
\caption{Comparison with state-of-the-art methods on the in-domain and cross-domain test sets. The best results are in \textbf{bold}.}
\label{tab:tracking_results}
\end{table*}
After the contextual information is effectively aggregated, we perform an adaptive channel-wise feature recalibration to accentuate the feature channels crucial for small object recognition and suppress potential background noise. We capture local cross-channel interaction information via a one-dimensional convolution ($\text{Conv}^{1D}_k$) with a kernel size of $k$, which is adaptively determined by a mapping function $\psi$ based on the channel dimension $C$:
\begin{equation}
\begin{aligned}
\mathbf{V'}^{(3)} &= \mathbf{w} \odot V_{ac}^{(3)}, \\
\mathbf{w} &= \sigma \left( \text{Conv}^{1D}_{k} \left( \text{GAP}(V_{ac}) \right) \right), \\
k &= \left| \frac{\log_2(C)+b}{\gamma} \right|_{\text{odd}},
\end{aligned}
\end{equation}
where $\gamma$ and $b$ are the hyperparameters and set to $\gamma = 2$ and $b = 1$, respectively. $|\cdot|_{\text{odd}}$ denotes the  nearest odd integer. GAP is global average pooling, and $\sigma$ is a Sigmoid function. 

Finally, the refined multi-scale visual features $\mathbf{F'}_v=\{ \mathbf{V'}^{(l)} \}_{l=1}^{L_s} $, refined by the encoder, are fed into the decoder with object queries to learn the target representation $D_t$. Furthermore, we employ a Semantic Guidance Module to perform semantically target-aware. Its process is as follows:
\begin{equation}
\begin{aligned}
    Q'_{\text{det}} &= Q_{\text{det}} + \operatorname{CrossAttn}(Q_{\text{det}}, T_{w}, T_{w}), \\
    D_t &= \operatorname{Decoder}(\mathbf{F'}_v, \operatorname{Concat}(Q_{\text{tra}}, Q'_{\text{det}})).
\end{aligned}
\end{equation}
\subsection{Loss Functions}
To train the tracker, the loss is computed through a linear combination of four specialized loss terms:
\begin{equation}\mathcal{L} = \lambda_{cls} \mathcal{L}_{cls} + \lambda_{L_1} \mathcal{L}_{L_1} + \lambda_{giou} \mathcal{L}_{giou} + \lambda_{ref} \mathcal{L}_{ref}.
\end{equation}
where, the constituent losses $\mathcal{L}_{cls}$, $\mathcal{L}_{L_1}$, and $\mathcal{L}_{giou}$ correspond to the focal loss, L1 loss, and GIoU loss. Each term is scaled by a corresponding hyperparameter $\lambda$, which controls its relative importance during the training process.

\section{Experiments}
\subsection{Implementation Details}
The main architectural settings follow those in~\cite{refer-kitti}. The entire training is deployed on 8 NVIDIA A100 GPUs with a batch size of 1 for 100 epochs. We filtered the target bounding boxes by applying a score threshold of 0.5 and a referring matching score threshold $\beta_{ref}=0.4$. AerialMind utilizes 63 sequences from the VisDrone for the training set and the remaining 17 sequences for in-domain testing. Additionally, we select 13 representative sequences from the UAVDT to serve as the cross-domain test set. 

\subsection{Evaluation Metrics}
To evaluate the overall tracking performance on AerialMind, we adopt the standard Higher Order Tracking Accuracy $\text{HOTA}=\sqrt{\text{DetA}\cdot \text{AssA}}$ metric~\cite{luiten2021hota}. To facilitate deeper and more fine-grained diagnostic analysis of model performance, we introduce two attribute-based composite metrics: $\text{HOTA}_S$ (Scene-Robustness) and $\text{HOTA}_M$ (Motion-Resilience).
For $\text{HOTA}_{S}$, the set of attributes $\{A_{i}\}$ comprises Night, Occlusion, and Low Resolution; and the attributes of $\text{HOTA}_{M}$ comprise Viewpoint Change, Scale Variation, Fast Motion, and Rotation. The general formula of attribute-based metrics is:
$\text{HOTA}_A=\sqrt[N]{\prod_{i=1}^N\text{HOTA}_{A_i}}$, $N$ denotes the number of attributes included. 
\subsection{Quantitative Results}
We conduct extensive comparisons with state-of-the-art RMOT methods (MOTRV2~\cite{motrv2}, TransRMOT~\cite{refer-kitti}, TempRMOT~\cite{refer-kittiv2}, CDRMT~\cite{cdrmt}, MGLT~\cite{refer-bdd}). The detailed results are presented in Table~\ref{tab:tracking_results}.
\begin{figure}[t]
    \includegraphics[width=0.9\linewidth]{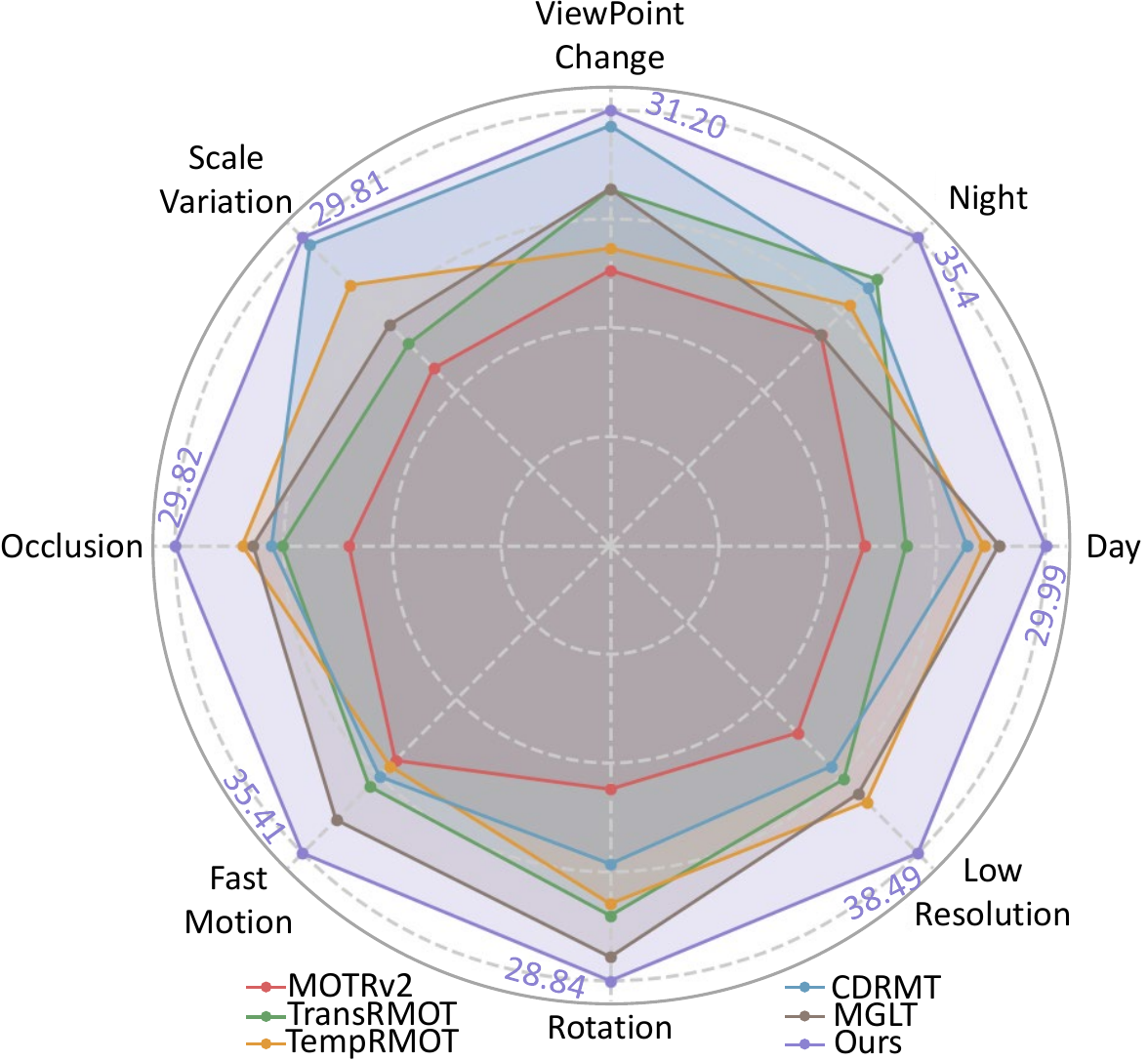}
    \caption{Comparison with state-of-the-art models in In-domain Evaluation with different attributes.}
    \label{fig:attri}
\end{figure}

\textbf{In-domain Evaluation.} HETrack demonstrates state-of-the-art performance, achieving a HOTA score of 31.46\%, significantly surpassing other leading end-to-end methods. Crucially, HETrack shows a pronounced advantage in our proposed attribute-based metric (HOTA$_S$ 34.37\%, HOTA$_M$ 31.12\%).  This superiority is further substantiated by a detailed attribute-based analysis, as visualized in Fig~\ref{fig:attri}. The results reveal that HETrack achieves the highest performance across all challenging attributes, and establishes a particularly significant lead in scenarios involving Low Resolution (38.49\%), Fast Motion (35.41\%), and Night (35.4\%) conditions. While HETrack enhances the ability for localizing these small-scale objects to improve overall detection accuracy (DetA 21.57\%), it leads to a marginal decrease in the average localization score (LocA 82.77\%).

\textbf{Cross-domain Evaluation.} To rigorously assess model generalization, we evaluate models on the cross-domain test set.  HETrack continues to outperform all other methods, not only achieving state-of-the-art results in core metrics like HOTA(31.60\%), DetA(21.35\%), and LocA(83.98\%), but also attaining the highest scores in our proposed attribute-based metrics, HOTA$_S$(27.53\%) and HOTA$_M$(31.93\%). 

An interesting phenomenon emerges from this evaluation: most methods, including HETrack, yield higher HOTA scores than their in-domain results. We posit that this counterintuitive result stems from the intrinsic disparity in scene complexity between the domains. Specifically, our training domain (VisDrone) features ten distinct object categories, fostering rich and complex semantic expressions. In contrast, the cross-domain test set (UAVDT) is predominantly limited to vehicle-only annotations, which significantly constrains the semantic space and simplifies the language grounding challenge. It also validates the distributional diversity and the value for pre-training of AerialMind.

\begin{table}[t]
\centering
\small
\setlength{\tabcolsep}{1pt} 
\begin{tabular}{@{}ccccccc@{}}
\toprule[1.5pt]
TransRMOT & TempRMOT & CDRMT & SKTrack & HFF-Track & HETrack \\
\midrule[1pt]
 31.00 &  35.04 & 31.99 & 35.29 & \textbf{36.18} & 35.40 \\
\bottomrule[1.5pt]
\end{tabular}
\caption{HOTA performance comparison of the Refer-KITTI-V2 dataset.}
\label{tab:hota_horizontal}
\end{table}

\begin{table}[t]
\begin{tabular*}{\columnwidth}{c|@{\extracolsep{\fill}}ccc}
    \toprule[1.5pt]
    Components & HOTA   & DetA   & AssA            \\
    \midrule
    w/o CFE \& SACR  & 26.41 & 16.43 & 42.80 \\ 
    w/o CFE     & 28.27 & 18.53 & 43.49    \\
    w/o SACR            & 29.89 & 19.86 & 45.34   \\
    HETrack (Ours)             & \textbf{31.46} & \textbf{21.57} & \textbf{46.23}    \\
    \bottomrule[1.5pt]
\end{tabular*}
\caption{Ablation studies of different components in HETrack. ``w/o" denotes components not used.}
\label{tab:abl_com}    
\end{table}
\textbf{Ground-level Evaluation.}
As shown in Table~\ref{tab:hota_horizontal}, we compare with state-of-the-art methods like SKTrack~\cite{refer-sktrack} and HFF-Track~\cite{zhao2025hff} on the complex expressions ground-level referring dataset Refer-KITTI-V2.  Our method also demonstrates competitive performance (35.40\% HOTA). It validates that our method provides universal benefits for referring understanding. 

\subsection{Qualitative Results}
\begin{figure}[t]
    \includegraphics[width=\linewidth]{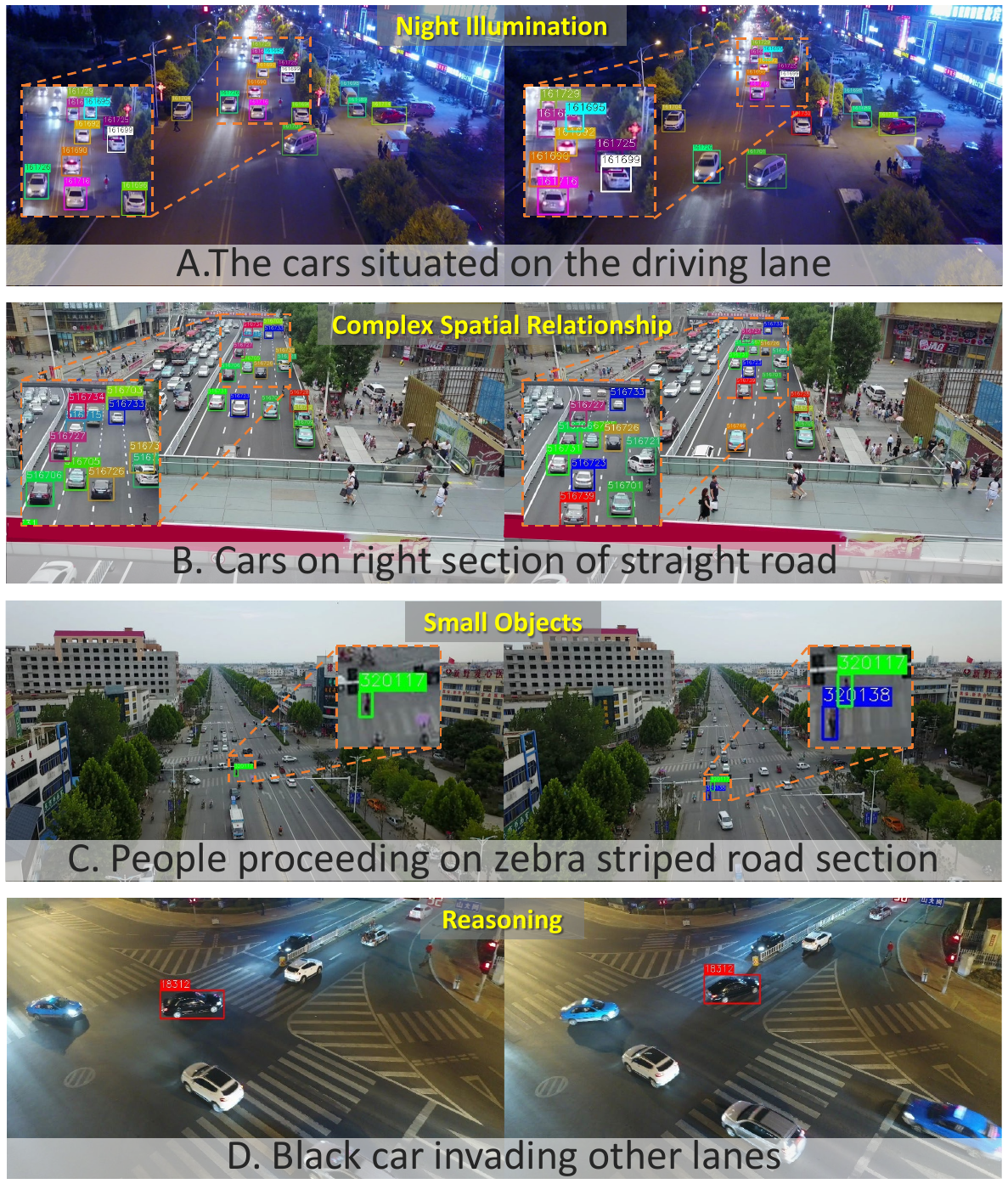}
    \caption{Qualitative examples on AerialMind. HETrack successfully tracks objects according to the expression.}
    \label{fig:vis}
\end{figure}

We visualize several representative examples in Figure~\ref{fig:vis}. HETrack successfully achieves precise detection and tracking of referent objects according to the given expressions in various challenging UAV scenarios, including night illumination, complex spatial relationships, and small objects. Most notably, Figure 6-D fully demonstrates the model's advanced reasoning capabilities for implicit descriptions. 
\subsection{Ablation Studies}
We systematically evaluate the contribution of our two main innovations (Table~\ref{tab:abl_com}). Our HETrack model achieves a state-of-the-art HOTA score of 31.46\%. When the SACR module is removed, the performance drops to 29.89\%, underscoring its critical role in enhancing small object perception. Removing the CFE module leads to a more significant performance degradation, with the HOTA score falling to 28.27\%, which highlights the importance of our synergistic vision-language fusion strategy. 
In Table~\ref{tab:abl_fus}, we compare different fusion strategies like feature concatenation, addition, and cross-attention with sentence-level features ($T_s$).

\begin{table}[t]
\setcellgapes{0pt} 
\makegapedcells 
\begin{tabular*}{\columnwidth}{c|@{\extracolsep{\fill}}ccc}
    \toprule[1.5pt]
    Fusion methods & HOTA  & DetA  & AssA         \\
    \midrule
    Concat        & 28.88 & 18.76 & 44.83   \\ 
    Add           & 30.39 & 19.95 & 46.65 \\ 
    Cross-Attn.($T_s$)   & 30.52 & 19.21 & \textbf{48.82}  \\
    Ours          & \textbf{31.46} & \textbf{21.57} & 46.23  \\
    \bottomrule[1.5pt]
\end{tabular*}
\caption{Ablation studies of Semantic Guidance Module.}
\label{tab:abl_fus}
\end{table}

\section{Conclusion}
In this work, we propose the first large-scale referring multi-object tracking dataset in UAV scenarios. It presents the unique challenges inherent in aerial viewpoints and introduces fine-grained attribute evaluation. Additionally, we develop a novel semi-automated collaborative agent-based framework that significantly enhances annotation efficiency and quality. Furthermore, we propose HETrack as a strong performance baseline. Extensive experiments validate the challenging nature of our dataset and the superior effectiveness of HETrack. We hope this work paves the way for future research in aerial language-guided perception.

\section{Acknowledgments}
This work was supported in part by the National Natural Science Foundation of China under Grant 62172246, in part by  Shandong Taishan Scholar Young Expert Project and Excellent Young Scientists Fund of Shandong Provincial Natural Science Foundation under Grant ZR2024YQ071, and in part by the Fundamental Research Funds for the Central Universities under Grant 22CX06037A, and in part by the Youth Innovation and Technology Support Plan of Colleges and Universities in Shandong Province under Grant 2021K1062, in part by the Criminal Inspection Key Laboratory of Sichuan Province under Grant 2024YB01, and in part by the Fundamental Research Funds for the Central Universities through the Youth Program under Grant 22CX06037A.

\bibliography{aaai2026}

\begin{thebibliography}{50}
\providecommand{\natexlab}[1]{#1}

\bibitem[{Chen et~al.(2025{\natexlab{a}})Chen, Lin, Zhong, Yao, and Li}]{refer-bdd}
Chen, J.; Lin, J.; Zhong, G.; Yao, Y.; and Li, Z. 2025{\natexlab{a}}.
\newblock Multi-granularity Localization Transformer with Collaborative Understanding for Referring Multi-Object Tracking.
\newblock \emph{IEEE Transactions on Instrumentation and Measurement}.

\bibitem[{Chen et~al.(2024{\natexlab{a}})Chen, Yu, Li, and Tao}]{delving}
Chen, S.; Yu, E.; Li, J.; and Tao, W. 2024{\natexlab{a}}.
\newblock Delving into the trajectory long-tail distribution for muti-object tracking.
\newblock In \emph{Proceedings of the IEEE/CVF Conference on Computer Vision and Pattern Recognition}, 19341--19351.

\bibitem[{Chen, Yu, and Tao(2025)}]{refer-cross}
Chen, S.; Yu, E.; and Tao, W. 2025.
\newblock Cross-view referring multi-object tracking.
\newblock In \emph{Proceedings of the AAAI Conference on Artificial Intelligence}.

\bibitem[{Chen et~al.(2025{\natexlab{b}})Chen, Yu, Yu, and Tao}]{reamot}
Chen, S.; Yu, Y.; Yu, E.; and Tao, W. 2025{\natexlab{b}}.
\newblock ReaMOT: A Benchmark and Framework for Reasoning-based Multi-Object Tracking.
\newblock \emph{arXiv preprint arXiv:2505.20381}.

\bibitem[{Chen et~al.(2025{\natexlab{c}})Chen, Jia, Lai, Wu, Xiao, Hu, and Yue}]{chen2025free}
Chen, W.; Jia, H.; Lai, S.; Wu, K.; Xiao, H.; Hu, L.; and Yue, Y. 2025{\natexlab{c}}.
\newblock Free-T2M: Frequency Enhanced Text-to-Motion Diffusion Model With Consistency Loss.
\newblock \emph{arXiv preprint arXiv:2501.18232}.

\bibitem[{Chen et~al.(2024{\natexlab{b}})Chen, Xiao, Zhang, Hu, Wang, Liu, and Chen}]{chen2024sato}
Chen, W.; Xiao, H.; Zhang, E.; Hu, L.; Wang, L.; Liu, M.; and Chen, C. 2024{\natexlab{b}}.
\newblock Sato: Stable text-to-motion framework.
\newblock In \emph{Proceedings of the 32nd ACM International Conference on Multimedia}, 6989--6997.

\bibitem[{Chen et~al.(2025{\natexlab{d}})Chen, Yu, Haozhe, Yuan, Huang, Tian, Lai, Xiao, Zhang, Wang et~al.}]{chen2025ant}
Chen, W.; Yu, K.; Haozhe, J.; Yuan, K.; Huang, Z.; Tian, B.; Lai, S.; Xiao, H.; Zhang, E.; Wang, L.; et~al. 2025{\natexlab{d}}.
\newblock ANT: Adaptive Neural Temporal-Aware Text-to-Motion Model.
\newblock In \emph{Proceedings of the 33rd ACM International Conference on Multimedia}, 9852--9861.

\bibitem[{Ding et~al.(2023)Ding, Liu, He, Jiang, Torr, and Bai}]{ding2023mose}
Ding, H.; Liu, C.; He, S.; Jiang, X.; Torr, P.~H.; and Bai, S. 2023.
\newblock MOSE: A new dataset for video object segmentation in complex scenes.
\newblock In \emph{Proceedings of the IEEE/CVF international conference on computer vision}, 20224--20234.

\bibitem[{Ding et~al.(2025{\natexlab{a}})Ding, Liu, He, Ying, Jiang, Loy, and Jiang}]{ding2025mevis}
Ding, H.; Liu, C.; He, S.; Ying, K.; Jiang, X.; Loy, C.~C.; and Jiang, Y.-G. 2025{\natexlab{a}}.
\newblock MeViS: A multi-modal dataset for referring motion expression video segmentation.
\newblock \emph{IEEE Transactions on Pattern Analysis and Machine Intelligence}.

\bibitem[{Ding et~al.(2022{\natexlab{a}})Ding, Liu, Wang, and Jiang}]{ding2022vlt}
Ding, H.; Liu, C.; Wang, S.; and Jiang, X. 2022{\natexlab{a}}.
\newblock VLT: Vision-language transformer and query generation for referring segmentation.
\newblock \emph{IEEE Transactions on Pattern Analysis and Machine Intelligence}, 45(6): 7900--7916.

\bibitem[{Ding et~al.(2025{\natexlab{b}})Ding, Ying, Liu, He, Jiang, Jiang, Torr, and Bai}]{MOSEv2}
Ding, H.; Ying, K.; Liu, C.; He, S.; Jiang, X.; Jiang, Y.-G.; Torr, P.~H.; and Bai, S. 2025{\natexlab{b}}.
\newblock {MOSEv2}: A More Challenging Dataset for Video Object Segmentation in Complex Scenes.
\newblock \emph{arXiv preprint arXiv:2508.05630}.

\bibitem[{Ding et~al.(2022{\natexlab{b}})Ding, Hui, Huang, Wei, Han, and Liu}]{video1}
Ding, Z.; Hui, T.; Huang, J.; Wei, X.; Han, J.; and Liu, S. 2022{\natexlab{b}}.
\newblock Language-bridged spatial-temporal interaction for referring video object segmentation.
\newblock In \emph{Proceedings of the IEEE/CVF Conference on Computer Vision and Pattern Recognition}.

\bibitem[{Du et~al.(2018)Du, Qi, Yu, Yang, Duan, Li, Zhang, Huang, and Tian}]{UAVDT}
Du, D.; Qi, Y.; Yu, H.; Yang, Y.; Duan, K.; Li, G.; Zhang, W.; Huang, Q.; and Tian, Q. 2018.
\newblock The unmanned aerial vehicle benchmark: Object detection and tracking.
\newblock In \emph{Proceedings of the European conference on computer vision (ECCV)}.

\bibitem[{Du et~al.(2019)Du, Zhu, Wen, Bian, Lin, Hu, Peng, Zheng, Wang, Zhang et~al.}]{visdrone}
Du, D.; Zhu, P.; Wen, L.; Bian, X.; Lin, H.; Hu, Q.; Peng, T.; Zheng, J.; Wang, X.; Zhang, Y.; et~al. 2019.
\newblock VisDrone-DET2019: The vision meets drone object detection in image challenge results.
\newblock In \emph{Proceedings of the IEEE/CVF international conference on computer vision workshops}.

\bibitem[{Du et~al.(2024)Du, Lei, Zhao, and Su}]{refer-dance}
Du, Y.; Lei, C.; Zhao, Z.; and Su, F. 2024.
\newblock ikun: Speak to trackers without retraining.
\newblock In \emph{Proceedings of the IEEE/CVF Conference on Computer Vision and Pattern Recognition}.

\bibitem[{Guan et~al.(2025{\natexlab{a}})Guan, Jia, Yao, Yang, Xu, Purwanto, Zhu, Man, Lim, Smith et~al.}]{guan2025watervg}
Guan, R.; Jia, L.; Yao, S.; Yang, F.; Xu, S.; Purwanto, E.; Zhu, X.; Man, K.~L.; Lim, E.~G.; Smith, J.; et~al. 2025{\natexlab{a}}.
\newblock Watervg: Waterway visual grounding based on text-guided vision and mmwave radar.
\newblock \emph{IEEE Transactions on Intelligent Transportation Systems}.

\bibitem[{Guan et~al.(2024)Guan, Liu, Jia, Zhao, Yao, Zhu, Man, Lim, Smith, and Yue}]{guan2024nanomvg}
Guan, R.; Liu, J.; Jia, L.; Zhao, H.; Yao, S.; Zhu, X.; Man, K.~L.; Lim, E.~G.; Smith, J.; and Yue, Y. 2024.
\newblock NanoMVG: USV-centric low-power multi-task visual grounding based on prompt-guided camera and 4D mmWave radar.
\newblock \emph{arXiv preprint arXiv:2408.17207}.

\bibitem[{Guan et~al.(2025{\natexlab{b}})Guan, Ouyang, Xu, Liang, Dai, Sun, Gao, Lai, Yao, Hu et~al.}]{guan2025yu}
Guan, R.; Ouyang, N.; Xu, T.; Liang, S.; Dai, W.; Sun, Y.; Gao, S.; Lai, S.; Yao, S.; Hu, X.; et~al. 2025{\natexlab{b}}.
\newblock Da Yu: Towards USV-Based Image Captioning for Waterway Surveillance and Scene Understanding.
\newblock \emph{arXiv preprint arXiv:2506.19288}.

\bibitem[{Guan et~al.(2025{\natexlab{c}})Guan, Zhang, Ouyang, Liu, Man, Cai, Xu, Smith, Lim, Yue et~al.}]{guan2025talk2radar}
Guan, R.; Zhang, R.; Ouyang, N.; Liu, J.; Man, K.~L.; Cai, X.; Xu, M.; Smith, J.; Lim, E.~G.; Yue, Y.; et~al. 2025{\natexlab{c}}.
\newblock Talk2radar: Bridging natural language with 4d mmwave radar for 3d referring expression comprehension.
\newblock In \emph{2025 IEEE International Conference on Robotics and Automation (ICRA)}, 10884--10891. IEEE.

\bibitem[{He et~al.(2016)He, Zhang, Ren, and Sun}]{resnet}
He, K.; Zhang, X.; Ren, S.; and Sun, J. 2016.
\newblock Deep residual learning for image recognition.
\newblock In \emph{Proceedings of the IEEE conference on computer vision and pattern recognition}.

\bibitem[{Hui et~al.(2021)Hui, Huang, Liu, Ding, Li, Wang, Han, and Wang}]{video2}
Hui, T.; Huang, S.; Liu, S.; Ding, Z.; Li, G.; Wang, W.; Han, J.; and Wang, F. 2021.
\newblock Collaborative spatial-temporal modeling for language-queried video actor segmentation.
\newblock In \emph{Proceedings of the IEEE/CVF Conference on Computer Vision and Pattern Recognition}.

\bibitem[{Khoreva, Rohrbach, and Schiele(2019)}]{refer-davis}
Khoreva, A.; Rohrbach, A.; and Schiele, B. 2019.
\newblock Video object segmentation with language referring expressions.
\newblock In \emph{Computer Vision--ACCV 2018: 14th Asian Conference on Computer Vision, Perth, Australia, December 2--6, 2018, Revised Selected Papers, Part IV 14}. Springer.

\bibitem[{Lai et~al.(2023)Lai, Hu, Wang, Berti-Equille, and Wang}]{lai2023faithful}
Lai, S.; Hu, L.; Wang, J.; Berti-Equille, L.; and Wang, D. 2023.
\newblock Faithful vision-language interpretation via concept bottleneck models.
\newblock In \emph{The Twelfth International Conference on Learning Representations}.

\bibitem[{Lai et~al.(2025)Lai, Liao, Hu, Yang, Chen, Xiao, Tang, Liao, and Yue}]{lai2025learning}
Lai, S.; Liao, M.; Hu, Z.; Yang, J.; Chen, W.; Xiao, H.; Tang, J.; Liao, H.; and Yue, Y. 2025.
\newblock Learning New Concepts, Remembering the Old: Continual Learning for Multimodal Concept Bottleneck Models.
\newblock In \emph{Proceedings of the 33rd ACM International Conference on Multimedia}, 12314--12322.

\bibitem[{Li et~al.(2025{\natexlab{a}})Li, Liu, Liu, Fan, and Zhang}]{lamot}
Li, Y.; Liu, X.; Liu, L.; Fan, H.; and Zhang, L. 2025{\natexlab{a}}.
\newblock Lamot: Language-guided multi-object tracking.
\newblock In \emph{2025 IEEE International Conference on Robotics and Automation (ICRA)}, 6816--6822. IEEE.

\bibitem[{Li et~al.(2025{\natexlab{b}})Li, Zhou, Qin, and Wang}]{refer-sktrack}
Li, Y.; Zhou, S.; Qin, Z.; and Wang, L. 2025{\natexlab{b}}.
\newblock Visual-Linguistic Feature Alignment with Semantic and Kinematic Guidance for Referring Multi-Object Tracking.
\newblock \emph{IEEE Transactions on Multimedia}.

\bibitem[{Liang et~al.(2023)Liang, Wang, Zhou, Miao, Luo, and Yang}]{onestage1}
Liang, C.; Wang, W.; Zhou, T.; Miao, J.; Luo, Y.; and Yang, Y. 2023.
\newblock Local-global context aware transformer for language-guided video segmentation.
\newblock \emph{IEEE Transactions on Pattern Analysis and Machine Intelligence}.

\bibitem[{Liang et~al.(2025)Liang, Guan, Lian, Liu, Sun, Wu, Yue, Ding, and Xiong}]{cdrmt}
Liang, S.; Guan, R.; Lian, W.; Liu, D.; Sun, X.; Wu, D.; Yue, Y.; Ding, W.; and Xiong, H. 2025.
\newblock Cognitive Disentanglement for Referring Multi-Object Tracking.
\newblock \emph{Information Fusion}.

\bibitem[{Liao et~al.(2020)Liao, Liu, Li, Wang, Chen, Qian, and Li}]{onestage2}
Liao, Y.; Liu, S.; Li, G.; Wang, F.; Chen, Y.; Qian, C.; and Li, B. 2020.
\newblock A real-time cross-modality correlation filtering method for referring expression comprehension.
\newblock In \emph{Proceedings of the IEEE/CVF Conference on Computer Vision and Pattern Recognition}.

\bibitem[{Lindeberg(2013)}]{lindeberg2013scale}
Lindeberg, T. 2013.
\newblock \emph{Scale-space theory in computer vision}, volume 256.
\newblock Springer Science \& Business Media.

\bibitem[{Liu et~al.(2025)Liu, Chen, Wang, Tang, Zhang, Yan, Wang, Li, and Zhao}]{liu2025aerialvg}
Liu, J.; Chen, Q.; Wang, Z.; Tang, Y.; Zhang, Y.; Yan, C.; Wang, D.; Li, X.; and Zhao, B. 2025.
\newblock AerialVG: A Challenging Benchmark for Aerial Visual Grounding by Exploring Positional Relations.
\newblock \emph{arXiv preprint arXiv:2504.07836}.

\bibitem[{Liu et~al.(2019)Liu, Ott, Goyal, Du, Joshi, Chen, Levy, Lewis, Zettlemoyer, and Stoyanov}]{roberta}
Liu, Y.; Ott, M.; Goyal, N.; Du, J.; Joshi, M.; Chen, D.; Levy, O.; Lewis, M.; Zettlemoyer, L.; and Stoyanov, V. 2019.
\newblock Roberta: A robustly optimized bert pretraining approach.
\newblock \emph{arXiv preprint arXiv:1907.11692}.

\bibitem[{Loshchilov and Hutter(2019)}]{adaw1}
Loshchilov, I.; and Hutter, F. 2019.
\newblock Decoupled Weight Decay Regularization.
\newblock In \emph{International Conference on Learning Representations}.

\bibitem[{Luiten et~al.(2021)Luiten, Osep, Dendorfer, Torr, Geiger, Leal-Taix{\'e}, and Leibe}]{luiten2021hota}
Luiten, J.; Osep, A.; Dendorfer, P.; Torr, P.; Geiger, A.; Leal-Taix{\'e}, L.; and Leibe, B. 2021.
\newblock Hota: A higher order metric for evaluating multi-object tracking.
\newblock \emph{International journal of computer vision}.

\bibitem[{Luo et~al.(2020)Luo, Zhou, Sun, Cao, Wu, Deng, and Ji}]{onestage3}
Luo, G.; Zhou, Y.; Sun, X.; Cao, L.; Wu, C.; Deng, C.; and Ji, R. 2020.
\newblock Multi-task collaborative network for joint referring expression comprehension and segmentation.
\newblock In \emph{Proceedings of the IEEE/CVF Conference on computer vision and pattern recognition}.

\bibitem[{Luo and Shakhnarovich(2017)}]{early2}
Luo, R.; and Shakhnarovich, G. 2017.
\newblock Comprehension-guided referring expressions.
\newblock In \emph{Proceedings of the IEEE Conference on Computer Vision and Pattern Recognition}.

\bibitem[{Ma et~al.(2024)Ma, Yang, Cui, Zhao, Su, Liu, and Wang}]{mls}
Ma, Z.; Yang, S.; Cui, Z.; Zhao, Z.; Su, F.; Liu, D.; and Wang, J. 2024.
\newblock Mls-track: Multilevel semantic interaction in rmot.
\newblock \emph{arXiv preprint arXiv:2404.12031}.

\bibitem[{Seo, Lee, and Han(2020)}]{refer-youtube}
Seo, S.; Lee, J.-Y.; and Han, B. 2020.
\newblock Urvos: Unified referring video object segmentation network with a large-scale benchmark.
\newblock In \emph{Computer Vision--ECCV 2020: 16th European Conference, Glasgow, UK, August 23--28, 2020, Proceedings, Part XV 16}. Springer.

\bibitem[{Song et~al.(2022)Song, Song, Yang, and Chen}]{song2022improving}
Song, M.; Song, W.; Yang, G.; and Chen, C. 2022.
\newblock Improving RGB-D salient object detection via modality-aware decoder.
\newblock \emph{IEEE Transactions on Image Processing}, 31: 6124--6138.

\bibitem[{Sun et~al.(2020)Sun, Cao, Jiang, Zhang, Xie, Yuan, Wang, and Luo}]{onestage4}
Sun, P.; Cao, J.; Jiang, Y.; Zhang, R.; Xie, E.; Yuan, Z.; Wang, C.; and Luo, P. 2020.
\newblock Transtrack: Multiple object tracking with transformer.
\newblock \emph{arXiv preprint arXiv:2012.15460}.

\bibitem[{Sun et~al.(2025)Sun, Liu, Zhu, Gu, Zou, Liu, Xia, Du, and Xu}]{refdrone}
Sun, Z.; Liu, Y.; Zhu, H.; Gu, Y.; Zou, Y.; Liu, Z.; Xia, G.-S.; Du, B.; and Xu, Y. 2025.
\newblock RefDrone: A Challenging Benchmark for Referring Expression Comprehension in Drone Scenes.
\newblock \emph{arXiv preprint arXiv:2502.00392}.

\bibitem[{Wang et~al.(2024)Wang, Chen, Hao, Qin, and Fan}]{wang2024windb}
Wang, G.; Chen, C.; Hao, A.; Qin, H.; and Fan, D.-P. 2024.
\newblock Windb: hmd-free and distortion-free panoptic video fixation learning.
\newblock \emph{IEEE Transactions on Pattern Analysis and Machine Intelligence}.

\bibitem[{Wang et~al.(2025)Wang, Yang, Liao, Zheng, Dai, Li, Liu et~al.}]{wang2025uav}
Wang, X.; Yang, D.; Liao, Y.; Zheng, W.; Dai, B.; Li, H.; Liu, S.; et~al. 2025.
\newblock UAV-Flow Colosseo: A Real-World Benchmark for Flying-on-a-Word UAV Imitation Learning.
\newblock \emph{arXiv preprint arXiv:2505.15725}.

\bibitem[{Wu et~al.(2022)Wu, Dong, Shao, and Shen}]{video3}
Wu, D.; Dong, X.; Shao, L.; and Shen, J. 2022.
\newblock Multi-level representation learning with semantic alignment for referring video object segmentation.
\newblock In \emph{Proceedings of the IEEE/CVF Conference on Computer Vision and Pattern Recognition}.

\bibitem[{Wu et~al.(2023)Wu, Han, Wang, Dong, Zhang, and Shen}]{refer-kitti}
Wu, D.; Han, W.; Wang, T.; Dong, X.; Zhang, X.; and Shen, J. 2023.
\newblock Referring multi-object tracking.
\newblock In \emph{Proceedings of the IEEE/CVF conference on computer vision and pattern recognition}.

\bibitem[{Yu et~al.(2016)Yu, Poirson, Yang, Berg, and Berg}]{refcoco}
Yu, L.; Poirson, P.; Yang, S.; Berg, A.~C.; and Berg, T.~L. 2016.
\newblock Modeling context in referring expressions.
\newblock In \emph{Computer Vision--ECCV 2016: 14th European Conference, Amsterdam, The Netherlands, October 11-14, 2016, Proceedings, Part II 14}. Springer.

\bibitem[{Zhang, Wang, and Zhang(2023)}]{motrv2}
Zhang, Y.; Wang, T.; and Zhang, X. 2023.
\newblock Motrv2: Bootstrapping end-to-end multi-object tracking by pretrained object detectors.
\newblock In \emph{Proceedings of the IEEE/CVF conference on computer vision and pattern recognition}.

\bibitem[{Zhang et~al.(2024)Zhang, Wu, Han, and Dong}]{refer-kittiv2}
Zhang, Y.; Wu, D.; Han, W.; and Dong, X. 2024.
\newblock Bootstrapping Referring Multi-Object Tracking.
\newblock \emph{arXiv preprint arXiv:2406.05039}.

\bibitem[{Zhao et~al.(2025)Zhao, Hao, Zhang, Liu, Li, Sui, He, and Chen}]{zhao2025hff}
Zhao, Z.; Hao, Y.; Zhang, M.; Liu, Q.; Li, B.; Sui, D.; He, S.; and Chen, X. 2025.
\newblock HFF-Tracker: A Hierarchical Fine-grained Fusion Tracker for Referring Multi-Object Tracking.
\newblock In \emph{Proceedings of the AAAI Conference on Artificial Intelligence}.

\bibitem[{Zhou et~al.(2022)Zhou, Porikli, Crandall, Van~Gool, and Wang}]{early4}
Zhou, T.; Porikli, F.; Crandall, D.~J.; Van~Gool, L.; and Wang, W. 2022.
\newblock A survey on deep learning technique for video segmentation.
\newblock \emph{IEEE transactions on pattern analysis and machine intelligence}.

\end{thebibliography}
\appendix
\twocolumn[{
\renewcommand\twocolumn[1][]{#1}%
\maketitle
\begin{figure}[H]
\centering
\vspace{-21mm}
\hsize=\textwidth
    \includegraphics[width=2\columnwidth]{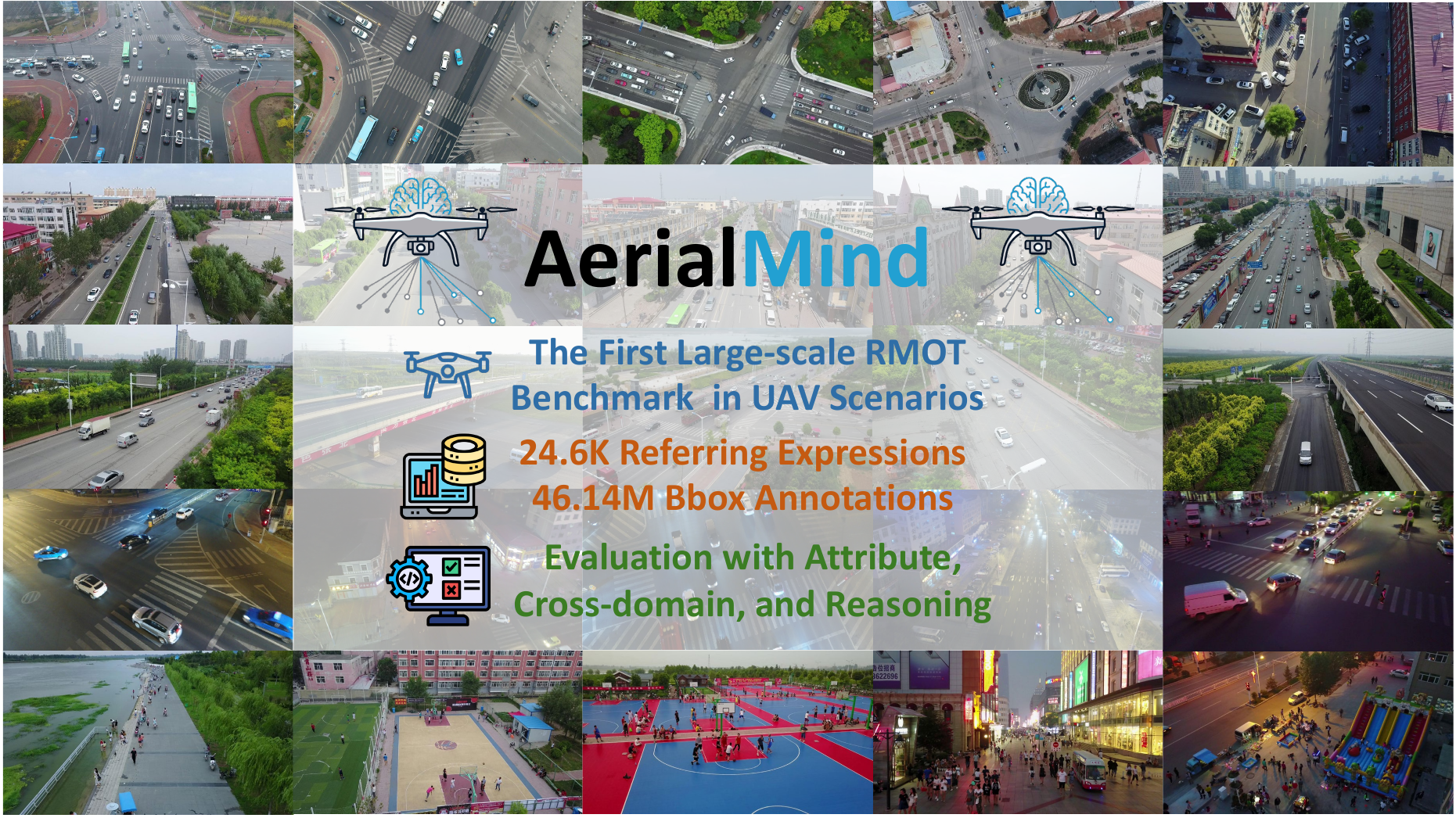}
    \caption{Overview of the AerialMind, including complex aerial scenarios, large-scale annotations, and diverse evaluation. 
    }
    \label{fig:aerialmind_overview}
\end{figure}
}]
\section{Appendix}
The following sections present comprehensive examinations of our AerialMind dataset, detailed methodology analysis, and extensive experimental evaluations to provide deeper insights into our contributions. \textbf{Section A} offers additional statistical analysis and visualizations of the AerialMind dataset, demonstrating its richness and complexity across multiple dimensions.  \textbf{Section B}  presents comprehensive implementation details, including detailed training strategies and hyperparameter configurations. It provides extensive experimental results comparing our method with state-of-the-art approaches on both AerialMind and existing benchmarks, including ablation studies and additional qualitative results. Finally, \textbf{Section C} discusses the broader implications of our work, limitations, and potential directions for future research in aerial language-guided perception. 
\section{A.AerialMind Benchmark}
As shown in Figure~\ref{fig:aerialmind_overview}, we demonstrate the diverse aerial scenarios that form the foundation of our benchmark. The collage of representative scenes, which range from complex urban intersections and highway interchanges to recreational facilities and commercial districts. It illustrates the unprecedented environmental diversity that distinguishes our dataset from existing ground-level benchmarks. As demonstrated in Table~\ref{tab:my_dataset_comparison}, our AerialMind dataset exhibits distinctive characteristics that establish it as a comprehensive benchmark for referring multi-object tracking in UAV Scenarios.
\subsection{Dataset Analysis and Statistics}
\begin{table*}[t!]
    \centering
    \setlength{\tabcolsep}{10pt} 
    \renewcommand{\arraystretch}{1.2} 
    \renewcommand\theadfont{\bfseries} 
    \begin{tabular}{@{}lcccccc@{}}
        \toprule[1.5pt]
        \thead{Dataset} & \thead{Source} & \thead{Frames} & \thead{Expressions Per \\ Sequence} & \thead{Distinct \\ Expressions} & \thead{Distinct \\ Instances} & \thead{Temporal Ratio \\ Per Expression} \\
        \midrule[1pt]
        Refer-KITTI   & CVPR$_{2023}$    & 6074   & 49.7  & 215  & 637  & 0.550   \\
        Refer-Dance    & CVPR$_{2024}$    & 67304  & 30.5  & 48   & 650  & \underline{0.939} \\
        Refer-BDD      & IEEE TIM$_{2025}$& 4610   & 92.2  & 1212 & 6246 & 0.538 \\
        Refer-UE-City  & arXiv$_{2024}$   & 6207   & 51    & --   & --   & 0.780   \\
        Refer-KITTIV2 & arXiv$_{2024}$   & 7938   & 464.6 & 7193 & 740  & 0.988  \\
        CRTrack          & AAAI$_{2025}$    & 82338 & 43.5  & 132  & --   & --     \\
        LaMOT* & ICRA$_{2025}$    & 27381  & 2.3   & 3    & 7889  & --     \\
        AerialMind       & Ours             & 48485  & 247.4 & 7601 & 8778 & 0.707  \\
        \bottomrule[1.5pt]
    \end{tabular}
    \caption{Comparison of referring multi-object tracking datasets. LaMOT* represents the UAV subset.}
    \label{tab:my_dataset_comparison}
\end{table*}
AerialMind contains 48,485 frames, which has a moderate scale compared to some existing datasets~\cite{refer-kitti,refer-bdd}. It reflects our deliberate focus on scene diversity rather than temporal redundancy. Unlike Refer-Dance~\cite{refer-dance} and CRMOT~\cite{refer-cross}, which achieve high frame counts through repetitive indoor scenarios (dance studios/stages) or multi-viewpoint captures of limited environments, our dataset encompasses over 70 distinct aerial scenarios, providing richer environmental diversity and visual complexity.

Our dataset achieves 247.4 expressions per sequence, ranking second among all benchmarks, which reflects our emphasis on sophisticated semantic interactions rather than simplistic lexical permutations~\cite{refer-kittiv2}.  Our commitment is to capturing the nuanced linguistic complexity inherent in real-world aerial operations.
Most significantly, AerialMind leads in both distinct expressions (7,601) and distinct instances (8,778), demonstrating unprecedented semantic richness and object diversity. This achievement validates our dataset's capacity to encompass the full spectrum of aerial referring scenarios, from fine-grained object discrimination to complex multi-target reasoning tasks. 

The temporal ratio per expression of 0.707 reveals another crucial design principle: our referring events are temporally dynamic rather than persistent throughout entire sequences. This characteristic reflects the realistic nature of aerial surveillance and monitoring tasks, where target objects frequently enter and exit the field of view due to UAV mobility and scene complexity. Unlike datasets with near-complete temporal coverage (e.g., Refer-KITTIV2~\cite{refer-kittiv2} at 0.988), our design captures the intermittent and event-driven nature of real-world aerial operations.

\begin{figure*}[!t]
	\centering
	\includegraphics[width=0.93\linewidth]{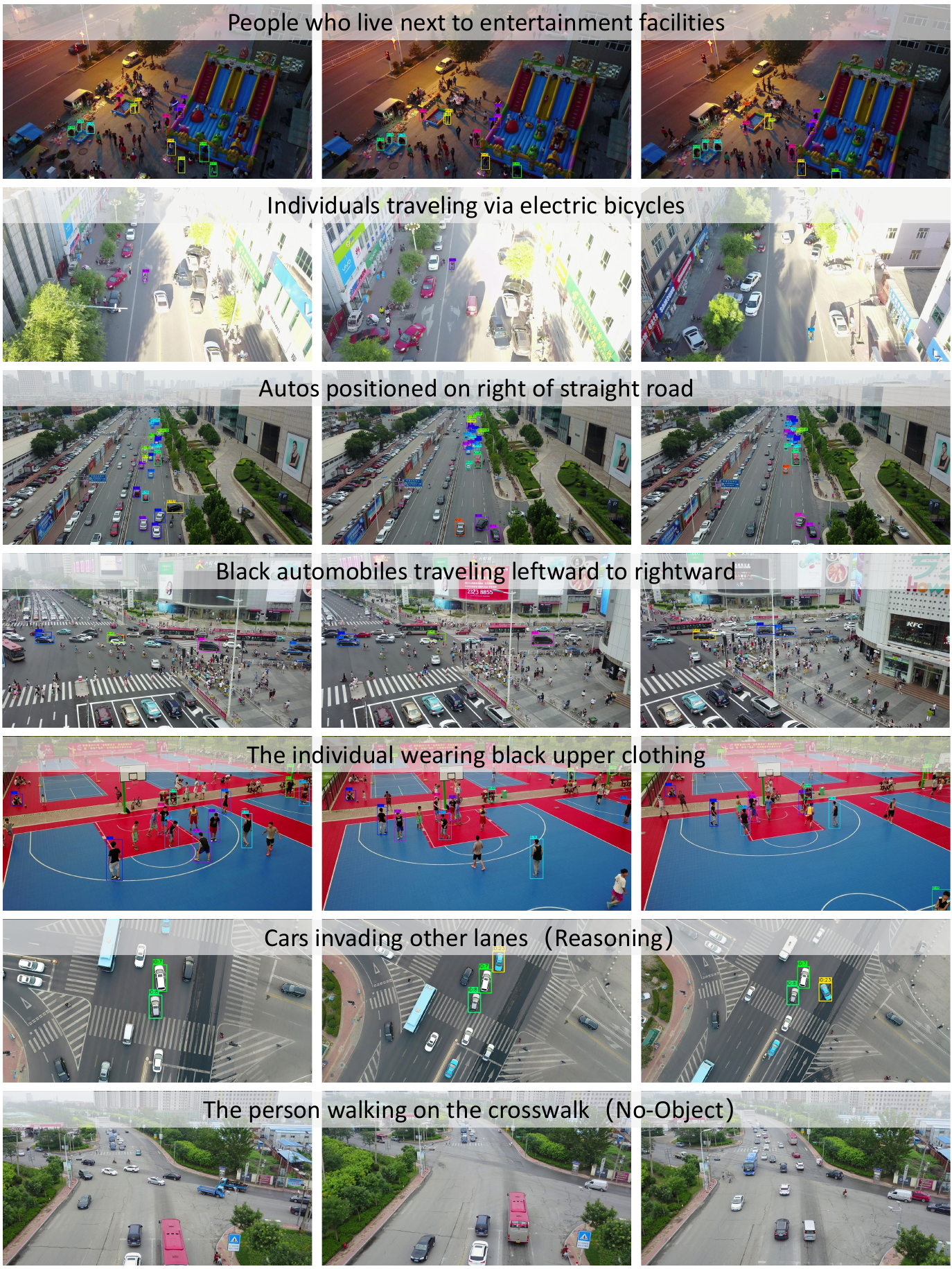}
	\caption{
		Representative examples from the AerialMind demonstrating diverse referring expressions and challenging scenarios.
	}
	\label{fig:dataset}
\end{figure*}

As shown in Figure~\ref{fig:dataset}, these examples demonstrate several key innovations: \textbf{(1) Complex spatial understanding}, exemplified by expressions like ``People who live next to entertainment facilities" and ``Autos positioned on right of straight road", which require understanding of complex spatial relationships from aerial perspectives; \textbf{(2) Fine-grained semantic discrimination}, such as distinguishing ``Individuals traveling via electric bicycles" from regular bicycles, demanding precise object categorization capabilities; \textbf{(3) Dynamic motion understanding}, as shown in ``Black automobiles traveling leftward to rightward", which integrates color attributes with directional motion analysis; \textbf{(4) Logical reasoning expressions}, including ``Cars invading other lanes", which requires logical inference about traffic violations rather than simple visual matching; \textbf{(5) Comprehensive negative sampling}, the no-target scenarios like ``The person walking on the crosswalk".

This combination of environmental diversity, semantic sophistication, and temporal dynamics positions AerialMind as a uniquely challenging benchmark that mirrors the complexity demands of practical aerial intelligence systems.

\subsection{Evaluation Metrics}
The design of our evaluation metrics framework is grounded in a fundamental understanding of the dual nature of challenges inherent in UAV-based referring multi-object tracking systems. Traditional RMOT~\cite{refer-kitti,mls} evaluation relies primarily on the standard HOTA metric~\cite{luiten2021hota}, which provides a balanced assessment of detection and association capabilities through the geometric mean $\text{HOTA}=\sqrt{\text{DetA}\cdot \text{AssA}}$. While HOTA offers valuable insights into overall system performance, it fails to capture the nuanced challenge taxonomy that distinguishes aerial platforms from ground-based systems. It limits our ability to conduct targeted diagnostic analysis and identify specific failure modes in complex UAV scenarios.

To address this limitation, we introduce a theoretically motivated decomposition of challenges into two fundamental categories that reflect the inherent characteristics of aerial operations. Our Scene-Robustness metric ($\text{HOTA}_S$) encompasses attributes that are intrinsic to the environmental context itself. Specifically, Night illumination conditions, Occlusion scenarios, and Low Resolution imagery. These attributes represent static environmental factors that exist independently of UAV behavior and would persist regardless of platform mobility or operational dynamics. 

\begin{table*}[t!]
\centering
\setlength{\tabcolsep}{12pt} 
\renewcommand{\arraystretch}{1.2} 
\begin{tabular}{@{}lccccccccc@{}}
\toprule[1.5pt]
\multicolumn{10}{c}{\textbf{Evaluation on Refer-KITTI-V2 Dataset}} \\
\midrule[1pt]
\thead{Method} & \thead{Source} & \thead{HOTA} & \thead{DetA} & \thead{AssA} & \thead{DetRe} & \thead{DetPr} & \thead{AssRe} & \thead{AssPr} & \thead{LocA} \\
\midrule[1pt]
FairMOT    & IJCV$_{2021}$  & 22.53 & 15.80 & 32.82          & 20.60          & 37.03          & 36.21          & 71.94          & 78.25 \\
ByteTrack  & ECCV$_{2022}$  & 24.59 & 16.78 & 36.63          & 22.60          & 36.18          & 41.00          & 69.63          & 78.00 \\
TransRMOT  & CVPR$_{2023}$  & 31.00 & 19.40 & 49.68          & 36.41          & 28.97          & 54.59          & 82.29          & 89.82 \\
iKUN       & CVPR$_{2024}$  & 10.32 & 2.17  & 49.77          & 2.36           & 19.75          & 58.48          & 68.64          & 74.56 \\
TempRMOT* & arXiv$_{2024}$ & 35.04 & 22.97 & \textbf{53.58} & 34.23          & 40.41          & \textbf{59.50} & 81.29          & \underline{90.07} \\
CDRMT      & INFFUS$_{2025}$& 31.99 & 20.37 & 50.35          & 26.40          & \textbf{46.26} & 53.40          & \textbf{85.90} & \textbf{90.36} \\
SKTrack    & TMM$_{2025}$   & 35.29 & 23.87 & 52.35          & \underline{39.97} & 36.48          & 57.45          & 84.23          & 88.89 \\
HFF-Track  & AAAI$_{2025}$  & \textbf{36.18} & \underline{24.64} & \underline{53.27} & 36.86          & \underline{41.83} & \underline{59.42} & 81.40          & 89.77 \\
HETrack    & Ours           & \underline{35.40} & \textbf{25.56} & 49.18          & \textbf{41.16} & 39.53          & 53.50          & \underline{85.51} & 89.80 \\
\bottomrule[1.5pt]
\end{tabular}
\caption{Comparison with state-of-the-art methods on the Refer-KITTI-V2 dataset. The best and second results are in \textbf{bold} and  \underline{underline}, respectively. * indicates that the result was obtained by performing inference using the official open source code.}
\label{tab:kitti_comparison}
\end{table*}
Conversely, our Motion-Resilience metric ($\text{HOTA}_M$) captures challenges that are directly correlated with UAV maneuverability and operational dynamics, including Viewpoint Change, Scale Variation, Fast Motion, and Camera Rotation. This categorization reflects the fundamental insight that UAV platforms introduce unique challenges through their mobility. Unlike static environmental challenges, these motion-induced factors are largely controllable through flight planning and operational procedures, making their separate evaluation crucial for understanding system limitations under different operational paradigms.

The mathematical formulation of our attributes evaluation metrics $\text{HOTA}_A=\sqrt[N]{\prod_{i=1}^N\text{HOTA}_{A_i}}$ employs the geometric mean to ensure that performance degradation in any constituent attribute significantly impacts the composite metric, thereby preventing compensation effects where strong performance in one attribute masks critical weaknesses in another. This approach is rooted in the fundamental principle that a robust model must demonstrate consistent reliability across all relevant challenge dimensions. Because operational failure in any single aspect can compromise mission effectiveness in real-world deployments.

\section{B.Additional Experimental Results}

\subsection{Implementation Details}
Our model employs ResNet50~\cite{resnet} as the visual backbone and RoBERTa~\cite{roberta} as the language encoder. Following established practices in referring multi-object tracking, the multi-scale feature maps extracted by the visual backbone undergo encoder-decoder processing for comprehensive cross-modal fusion and spatial relationship modeling. The remaining architectural configurations adhere to the framework established in~\cite{refer-kitti}. 

The training process spans 100 epochs using the AdamW~\cite{adaw1} optimizer with an initial learning rate of $1 \times 10^{-4}$. The learning rate undergoes decay by a factor of 10 at the $40^{th}$ epoch to ensure convergence stability. The loss function incorporates multiple weighted components: $\lambda_{cls}$, $\lambda_{L_1}$, $\lambda_{giou}$, and $\lambda_{align}$ are configured to 2, 5, 2, and 2, respectively, to balance classification accuracy, localization precision, and alignment quality. To accommodate the diverse linguistic expressions characteristic of aerial scenarios, we initialize the framework with 300 object queries. To ensure reproducibility and fair comparison across different runs, we maintain a fixed random seed throughout both training and testing phases. Our HETrack model with 51.4M trainable parameters achieves 15.6 FPS on a single RTX 4080 during inference.

\subsection{Theoretical Foundations and Design Motivations}
The architectural innovations in HETrack are grounded in rigorous theoretical foundations that address fundamental limitations in cross-modal representation learning and multi-scale object detection. This section elucidates the mathematical principles and theoretical motivations underlying our Co-evolutionary Fusion Encoder (CFE) and Scale Adaptive Contextual Refinement (SACR) modules.

\textbf{Co-evolutionary Fusion Encoder: Information-Theoretic Motivation.} Traditional fusion paradigms suffer from the information bottleneck problem~\cite{lai2023faithful,lai2025learning}, where cross-modal alignment is constrained by unidirectional information flow. Let $V$ and $L$ represent visual and linguistic feature distributions, respectively. Conventional approaches are limited by $I(V;L) \leq \min(H(V), H(L))$, where mutual information is bounded by the entropy of individual modalities. Our CFE addresses this fundamental limitation through bidirectional iteration.
It enables visual and linguistic representations to evolve synergistically rather than independently.

\textbf{Scale Adaptive Contextual Refinement: Multi-scale Information Theory.} The SACR module is motivated by multi-scale information processing theory~\cite{lindeberg2013scale}, which posits that optimal feature representation for small object detection requires principled information integration across multiple spatial scales. For UAV scenarios with extreme scale variations, the effective receptive ($ERF_{optimal}$) field must satisfy:
\begin{equation}
\begin{aligned}ERF_{optimal} = \arg\max_{r} \sum_{s \in \mathcal{S}} w_s \cdot MI(F_r^{(s)}, Y),
\end{aligned}   
\end{equation}
where $F_r^{(s)}$ represents features at scale $s$ with receptive field $r$, and $MI$ denotes mutual information with target labels $Y$. Our atrous convolution configuration with dilation rates $\{6,12,18\}$ approximates this optimal multi-scale integration while preserving spatial resolution.
The adaptive channel recalibration mechanism seeks to emphasize task-relevant features while reducing noise interference~\cite{guan2025watervg} for robust representation.
\begin{table*}[t!]
\centering
\setlength{\tabcolsep}{8.5pt} 
\renewcommand{\arraystretch}{1.2} 
\begin{tabular}{@{}lccccccccc@{}}
\toprule[1.5pt]
\multicolumn{10}{c}{\textbf{Attributes Evaluation on UAVDT test set}} \\
\midrule[1pt]
\thead{Method} & \thead{Source} & \thead{Day} & \thead{Night} & \thead{ViewPoint \\ Change} & \thead{Scale \\ Variation} & \thead{Occlusion} & \thead{Fast \\ Motion} & \thead{Rotation} & \thead{Low \\ Resolution} \\
\midrule[1pt]
MOTRv2     & CVPR$_{2023}$  & 22.07 & 20.42 & 26.25          & 19.86          & 28.53          & 28.10          & 26.01          & 23.28          \\
TransRMOT  & CVPR$_{2023}$  & \underline{28.90} & 23.00 & \underline{31.61} & \underline{22.28} & \underline{28.59} & 24.05          & 24.69          & 22.28          \\
TempRMOT   & arXiv$_{2024}$ & 28.27 & 25.48 & 29.33          & 20.65          & 26.15          & 24.40          & \textbf{39.64} & 20.08          \\
CDRMT      & INFFUS$_{2025}$& 26.64 & \underline{26.78} & 30.10          & 20.72          & \textbf{29.94} & 27.09          & 23.88          & 24.49          \\
MGLT       & IEEE TIM$_{2025}$ & 28.39 & 25.62 & 31.57          & 19.41          & 26.68          & \textbf{33.67} & 30.64          & \textbf{28.60} \\
HETrack    & Ours           & \textbf{32.18} & \textbf{30.16} & \textbf{33.57} & \textbf{26.23} & 25.80          & \underline{31.10} & \underline{37.96} & \underline{26.82} \\
\bottomrule[1.5pt]
\end{tabular}
\caption{Comparison of various models in Cross-domain Evaluation with different attributes. The best results are in \textbf{bold} and the second best are \underline{underlined}.}
\label{tab:scenario_comparison}
\end{table*}
\subsection{Quantitative Results}
\begin{figure*}[!t]
	\centering
	\includegraphics[width=\linewidth]{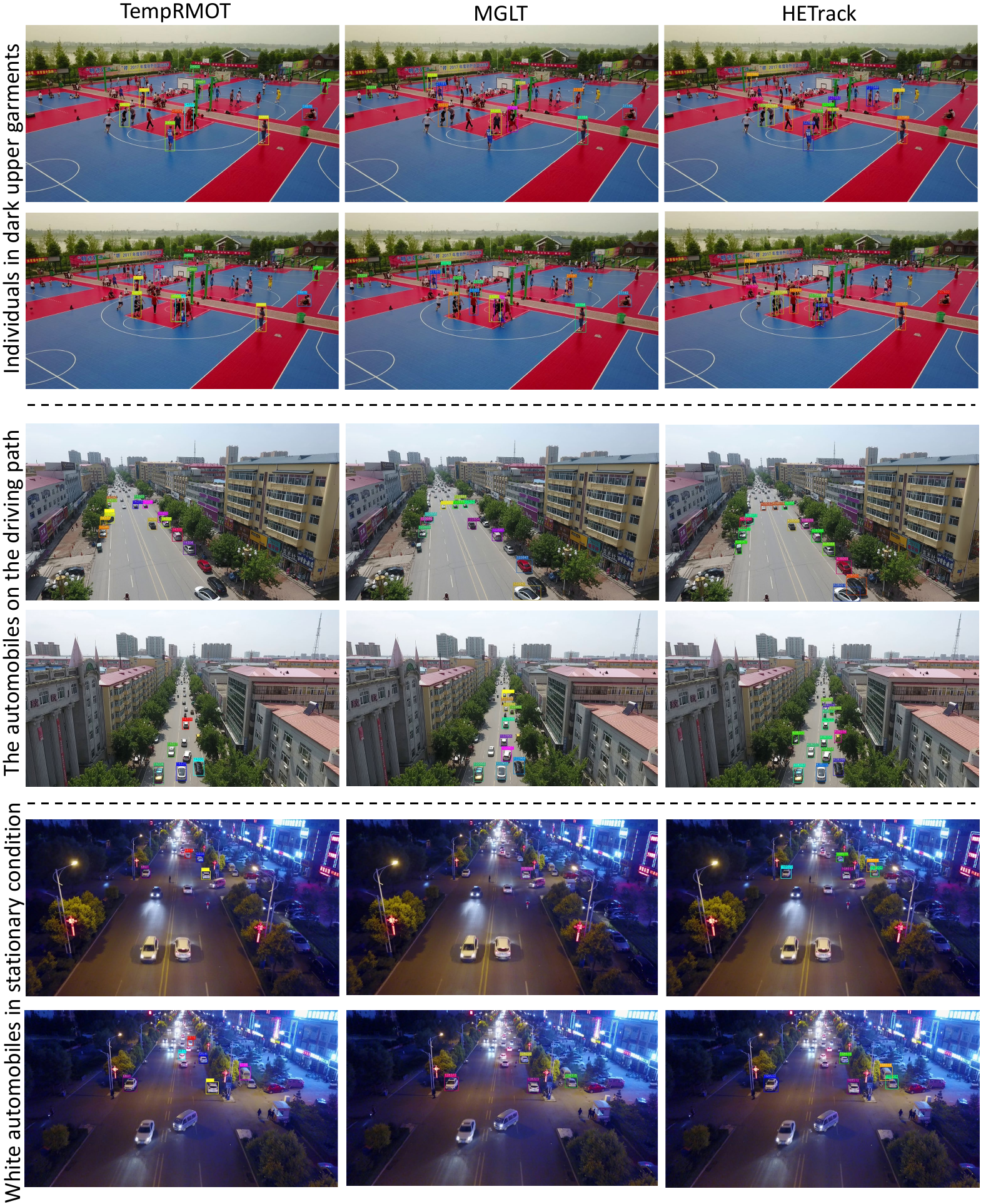}
	\caption{
		Qualitative comparison with the previous state-of-the-art methods on the In-domain evaluation of AerialMind.
	}
	\label{fig:indomain}
\end{figure*}
\begin{figure*}[!t]
	\centering
	\includegraphics[width=\linewidth]{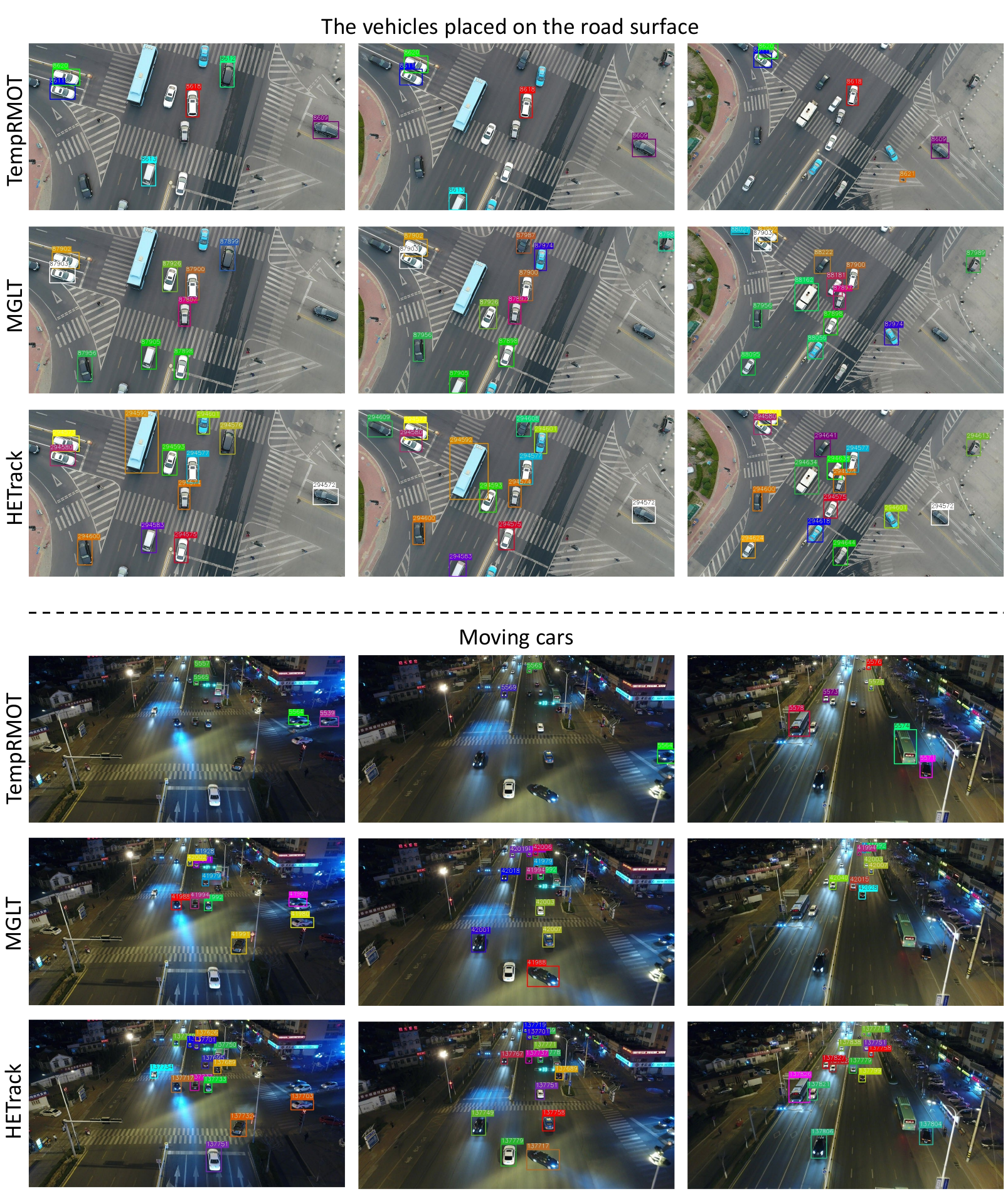}
	\caption{
		Qualitative comparison with the previous state-of-the-art methods on the Cross-domain evaluation of AerialMind.
	}
	\label{fig:crossdomain}
\end{figure*}
\begin{figure*}[!t]
	\centering
	\includegraphics[width=\linewidth]{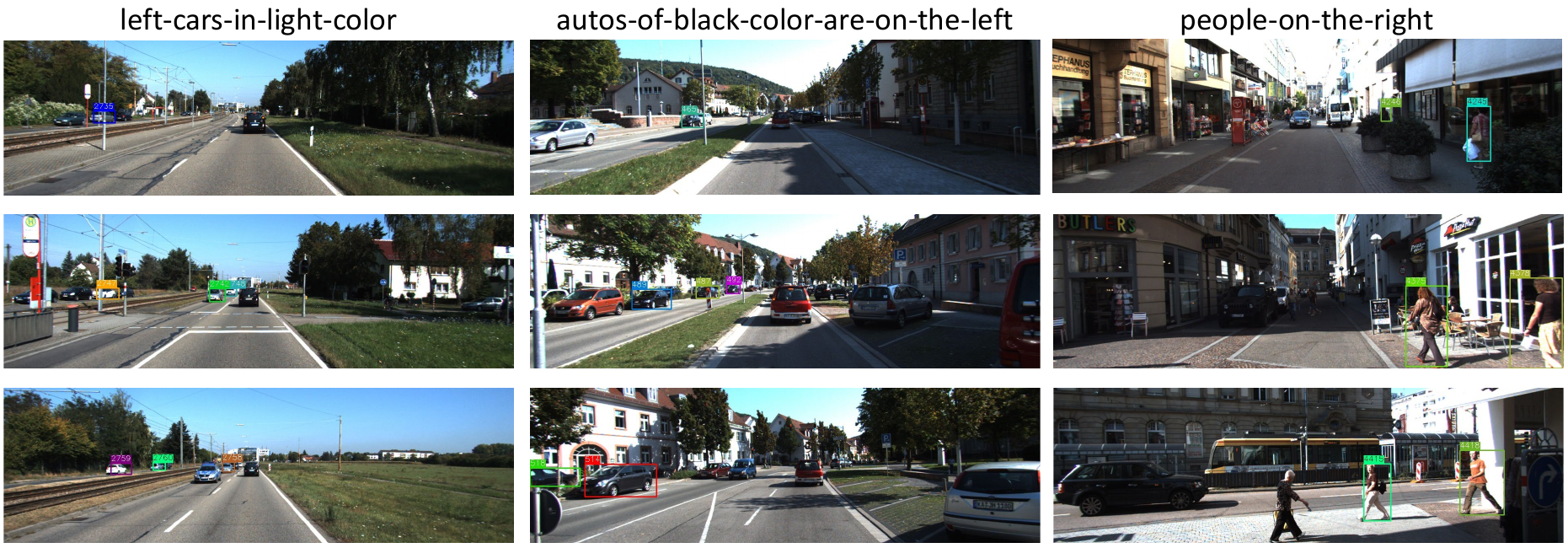}
	\caption{
		Qualitative results of our method on the Refer-KITTI-V2.
	}
	\label{fig:kitti}
\end{figure*}
\subsubsection{Evaluation on Refer-KITTI-V2 Dataset}
To further validate the generalizability and robustness of our proposed HETrack method, we conduct comprehensive evaluations on the Refer-KITTI-V2 dataset~\cite{refer-kittiv2}. It represents the most extensive ground-perspective RMOT benchmark featuring complex linguistic descriptions. It's a crucial evaluation for our method's adaptability across diverse scenarios and semantic complexities.

As demonstrated in Table~\ref{tab:kitti_comparison}, HETrack achieves competitive performance with a HOTA score of 35.40\%, ranking among the top-tier methods alongside SKTrack (35.29\%) and HFF-Track (36.18\%). Notably, our method demonstrates particular strengths in detection accuracy (DetA: 25.56\%) and detection recall (DetRe: 41.16\%), significantly outperforming recent methods such as CDRMT (DetA: 20.37\%) and SKTrack (DetA: 23.87\%). This superior detection capability directly validates the effectiveness of our Scale Adaptive Contextual Refinement (SACR) module, which, despite being originally designed for small object perception in aerial scenarios. It proves equally beneficial for ground-level object detection by enhancing contextual feature representation and reducing background noise.
More significantly, our method achieves remarkable improvements in critical detection metrics, with DetRe reaching 41.16\%—substantially higher than competing methods such as SKTrack (39.97\%) and TempRMOT (34.23\%). This enhancement in detection recall demonstrates the robust cross-modal alignment capabilities of our Co-evolutionary Fusion Encoder (CFE), which enables synergistic refinement of vision-language representations. The CFE's bidirectional fusion mechanism proves particularly effective in handling the complex linguistic expressions characteristic.

These results provide compelling evidence of our method's core innovations. They capture fundamental principles of referring multi-object tracking that generalize effectively across diverse visual domains. It demonstrates that HETrack constitutes a generalizable framework capable of adapting to different scene characteristics while maintaining its core strengths in cross-modal representation learning and contextual feature enhancement.
\subsubsection{Attributes evaluation on UAVDT}
To provide a comprehensive assessment of our method's cross-domain generalization capabilities, we present detailed attribute-based performance analysis on the UAVDT test set in Table~\ref{tab:scenario_comparison}. 

HETrack's consistent top-tier performance across seven out of eight challenging attributes, achieving four best and three second-best results. It provides compelling evidence of its superior cross-domain generalization capabilities (Table~\ref{tab:scenario_comparison}). Specifically, HETrack achieves state-of-the-art performance in the crucial attributes of Day, Night, ViewPoint Change, and Scale Variation. Furthermore, it demonstrates highly competitive results in Fast Motion, Rotation, and Low Resolution.  These results confirm that our innovations successfully capture domain-agnostic tracking principles and are not overfit to the training domain's data.
\subsection{Qualitative Results}
To provide comprehensive insights into the practical effectiveness of our proposed HETrack method, we present detailed qualitative comparisons with state-of-the-art approaches TempRMOT~\cite{refer-kittiv2} and MGLT~\cite{refer-bdd} across challenging scenarios in Figure~\ref{fig:indomain}-\ref{fig:kitti}. 
\subsubsection{In-domain Evaluation}
As shown in Figure~\ref{fig:indomain}, the basketball court scene presents a particularly demanding test case involving dense crowds, frequent occlusions, and rapid multi-directional movements. Both TempRMOT~\cite{refer-kittiv2} and MGLT~\cite{refer-bdd} exhibit failures in detecting ``individuals wearing dark upper garments" in this complex scene. In stark contrast, HETrack robustly detects and tracks all relevant individuals.

The second row showcases HETrack's exceptional capability in handling the dual challenges of small object detection and partial occlusion by environmental elements. While TempRMOT and MGLT fail to comprehensively identify automobiles along the driving path, missing numerous roadside vehicles and distant objects, HETrack demonstrates remarkable precision in detecting both prominently visible and partially occluded vehicles. 

The nighttime intersection scene presents a more sophisticated challenge, requiring accurate interpretation of motion states rather than visual appearance matching. The referring expression ``white automobiles in stationary condition" demands precise discrimination between stationary and moving vehicles of the same color category. TempRMOT~\cite{refer-kittiv2} demonstrates semantic confusion by incorrectly identifying moving white vehicles as targets, while MGLT~\cite{refer-bdd} suffers from incomplete detection, missing several legitimate stationary targets. HETrack achieves superior semantic precision by correctly identifying only the stationary white automobiles.

\subsubsection{Cross-domain Evaluation}
The cross-domain evaluation on UAVDT (Figure~\ref{fig:crossdomain}) demonstrates HETrack's superior generalization capabilities across diverse scenarios and challenging conditions. In the complex intersection scenario with ``vehicles placed on the road surface," both TempRMOT and MGLT exhibit significant detection failures, particularly missing small-scale vehicles in distant regions. HETrack demonstrates remarkable robustness, successfully detecting vehicles across various scales and maintaining stable associations throughout the temporal sequence.

The night ``moving cars" scenario reveals critical limitations in existing methods' motion understanding and low-light object detection capabilities. TempRMOT shows poor comprehension of motion states, failing to accurately distinguish moving vehicles from stationary ones, while both TempRMOT and MGLT struggle with detecting small-scale targets under challenging illumination conditions. In contrast, HETrack maintains comprehensive detection of small nighttime objects and demonstrates superior motion state interpretation, accurately identifying only the moving vehicles as specified by the referring expression.

\subsubsection{Ground-level Evaluation}
To further validate the generalizability of our approach across different viewing perspectives, we evaluate HETrack on ground-level scenarios from the Refer-KITTI-V2 dataset (Figure~\ref{fig:kitti}). Despite being primarily designed for aerial scenarios, our method demonstrates remarkable adaptability to ground-perspective referring tasks. Across diverse urban environments and different viewpoint conditions, HETrack achieves precise object identification according to the given referring expressions.

\begin{table}[t]
\begin{tabular*}{\columnwidth}{c|@{\extracolsep{\fill}}ccc}
    \toprule[1.5pt]
    Method & HOTA   & DetA   & AssA            \\
    \midrule
    0.3             & 27.01 & 18.41 & 40.04 \\ 
    0.4    & \textbf{31.46} & \textbf{21.57} & 46.23    \\
    0.5             & 30.72 & 20.32 & \textbf{46.83}   \\
    0.6             & 26.64 & 16.88 & 42.44    \\
    \bottomrule[1.5pt]
\end{tabular*}
\caption{Ablation studies on referring threshold $\beta_{ref}$.}
\label{tab:abl_ref}
\end{table}
\subsection{Ablation Study}
\subsubsection{Referring Threshold}
As demonstrated in Table~\ref{tab:abl_ref}, the model exhibits distinct performance patterns across the different thresholds. At the conservative threshold of $\beta_{ref}=0.3$, it enables the model to capture more potential matches but introduces noise through false positive associations, as evidenced by the relatively lower DetA score of 18.41\%.
The optimal performance is achieved at $\beta_{ref}=0.4$, where the model attains peak HOTA (31.46\%) and DetA (21.57\%) scores. This threshold represents an effective balance between detection sensitivity and association precision, enabling the model to maintain robust referring accuracy while maximizing overall tracking performance. 
Interestingly, increasing the threshold to $\beta_{ref}=0.5$ results in marginal degradation in overall performance (HOTA: 30.72\%) despite achieving the highest association accuracy (AssA: 46.83\%). This pattern indicates that while stricter thresholds improve the quality of established associations by filtering out ambiguous matches, they simultaneously reduce the system's ability to detect valid referring instances, as reflected in the decreased DetA score (20.32\%). 
The pronounced performance degradation at $\beta_{ref}=0.6$ (HOTA: 26.64\%, DetA: 16.88\%, AssA: 42.44\%) confirms that excessively strict thresholds severely limit the model's detection capabilities. At this threshold, the system becomes overly selective, potentially missing legitimate referring instances due to the inherent uncertainty in cross-modal matching.

\begin{table}[t]
\centering
\begin{tabular*}{\columnwidth}{c|@{\extracolsep{\fill}}ccc}
    \toprule[1.5pt]
    SACR Components & HOTA & DetA & AssA \\
    \midrule
    Only Atrous Conv & 29.70 & 19.43 & 45.81 \\
    Only Channel Recalibration & 29.13 & 18.73 & 45.58 \\
    Full SACR (Ours) & \textbf{31.46} & \textbf{21.57} & \textbf{46.23} \\
    \bottomrule[1.5pt]
\end{tabular*}
\caption{Ablation study on SACR module components.}
\label{tab:abl_sacr_components}
\end{table}
\subsubsection{Components of SACR module}
Table~\ref{tab:abl_sacr_components} provides a systematic decomposition of the SACR module's effectiveness, revealing the synergistic contributions of its constituent components for small object detection in aerial scenarios. 
When employing only the atrous convolution component, the model achieves moderate performance (HOTA: 29.70\%, DetA: 19.43\%), validating the importance of multi-scale contextual information capture for object localization. However, the limited performance improvement indicates that contextual information alone is insufficient for optimal object detection.
The isolated channel recalibration mechanism yields relatively lower performance (HOTA: 29.13\%, DetA: 18.73\%), clearly suggesting that channel attention without adequate contextual support struggles to identify discriminative features for small-scale targets. 
The superior performance of our complete SACR module (HOTA: 31.46\%, DetA: 21.57\%) validates the synergistic integration of multi-scale contextual refinement and adaptive channel recalibration.  It proves particularly valuable for effectively handling the diverse feature representations encountered across different scales in challenging  UAV scenarios.
\section{C.Discussion}
In this work, we introduced AerialMind, the first large-scale benchmark for Referring Multi-Object Tracking (RMOT) in UAV scenarios, alongside a strong baseline method, HETrack. Our contributions aim to bridge the significant gap between prevailing ground-level research and the unique, complex challenges posed by aerial platforms, thereby pushing the community towards developing more robust and versatile language-guided perception systems.

The primary significance of AerialMind lies in its establishment of a standardized evaluation platform for the aerial domain. By systematically incorporating challenges such as drastic scale variations, complex spatial relationships, and dynamic scenes, complemented by the first-of-its-kind attribute-level annotations in the RMOT field, our benchmark facilitates a more granular and insightful analysis of model capabilities. However, as a foundational step, we acknowledge its current limitations. The benchmark is constructed by extending existing public datasets, namely VisDrone and UAVDT. They inherited from these foundational datasets the presence of minor, pre-existing annotation errors. While we implemented a rigorous process to ensure the quality of our own annotations, it prevents the completely correct ground-truth labeling in a small subset of cases.  Our future work will involve augmenting the benchmark with self-captured data, featuring a wider array of scenarios and more intricate object interactions.

While HETrack demonstrates competitive performance across various scenarios, several limitations warrant attention for future development. First, the current architecture relies on traditional vision-language fusion paradigms without leveraging the advanced reasoning capabilities of large language models (LLMs).  Second, the computational overhead of our method presents challenges for real-time deployment on resource-constrained UAV platforms. The development of lightweight RMOT model variants represents a crucial direction for future research.
\end{document}